\documentclass[twocolumn]{article}
\usepackage{PRIMEarxiv}

\usepackage[utf8]{inputenc} 
\usepackage[T1]{fontenc}    
\usepackage{hyperref}       
\usepackage{url}            
\usepackage{booktabs}       
\usepackage{amsfonts}       
\usepackage{nicefrac}       
\usepackage{microtype}      
\usepackage{lipsum}
\usepackage{fancyhdr}       
\usepackage{graphicx}       
\graphicspath{{media/}}     
\usepackage{natbib}
\usepackage{amsmath}
\usepackage{amsthm}
\newtheorem{proposition}{Proposition}
\usepackage{comment}
\usepackage{adjustbox}
\usepackage{subcaption}
\usepackage{dsfont}
\hypersetup{
    colorlinks=false,
    pdfborder={0 0 0},
    pdfborderstyle={/S/U/W 0},
}

\newcommand{\harm}[1][\mathcal{D}]{\ensuremath{\mathrm{H}_{#1}[c_w]}}

\newcommand{\rate}[1][\mathcal{D}]{\ensuremath{\mathrm{r}_{#1}[c_w]}}
\newcommand{\ratet}[1][\mathcal{D}]{\ensuremath{\mathrm{r}_{#1}[c_w>t_{#1}]}}

\newcommand{\harmfn}[1][c_w]{\ensuremath{\mathrm{H}_{\mathcal{D}}[#1]}}
  
\title{Resource-constrained Fairness}

\author{ \href{https://orcid.org/0000-0003-3784-826X}{\includegraphics[scale=0.06]{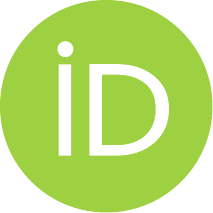}\hspace{1mm}Sofie~Goethals} \\
	University of Antwerp\\
	Antwerp, Belgium \\
	\texttt{sofie.goethals@uantwerpen.be} \\
    \And
        Eoin Delaney\\
	University of Oxford\\
	Oxford, United Kingdom \\
     \And
        Brent Mittelstadt\\
	University of Oxford\\
	Oxford, United Kingdom \\
     \And
        Chris Russell\\
	University of Oxford\\
	Oxford, United Kingdom \\
	}

\begin{document}

\twocolumn[
\maketitle
]

\begin{abstract}
Access to resources strongly constrains the decisions we make. While we might wish to offer every student a scholarship, or schedule every patient for follow-up meetings with a specialist, limited resources make this infeasible. 
When deploying machine learning systems, these resource constraints are typically enforced by adjusting the classifier threshold.
However, these finite resource limitations are disregarded by most existing tools for fair machine learning, which do not allow for the specification of resource limitations and do not remain fair when varying thresholds. This makes them ill-suited for real-world deployment.
Our research introduces the concept of ``\textit{resource-constrained fairness}" and quantifies the cost of fairness within these constraints. We demonstrate that the level of available resources significantly influences this cost, a factor overlooked in prior evaluations.
\end{abstract}

\keywords{Algorithmic Fairness \and Responsible AI in practice}

\section{Introduction}
Machine learning models are used to make decisions in many high-impact areas of our lives such as finance, justice, and healthcare \cite{mehrabi2021survey}.
Fair machine learning has emerged in response to the notion that simply making maximally accurate decisions is not enough and that training high-performance classifiers can result in both the transfer of existing biases from data to new decisions, as well as the introduction of new biases~\cite{wachter2020bias}.
Many studies that focus on improving fairness in machine learning overlook the practical limitations under which these models operate. For example, scenarios including university admissions, healthcare provision, and corporate hiring, are usually constrained by finite resources. Universities have a restricted quota of students to admit annually, healthcare facilities are bounded by available space and staff, and companies have a limited number of positions to fill. Even for banking, where in
principle banks can keep making loans providing enough people pay them
back, for any particular time period they will have a limited set of
resources and only be able to loan out a certain amount. 
In all these cases, \emph{once the available resources are fully used,} increasing the selection rate of disadvantaged groups must necessarily involve reducing selection from more advantaged groups.

\begin{figure}[t]
    \centering    \includegraphics[width=0.8\linewidth]{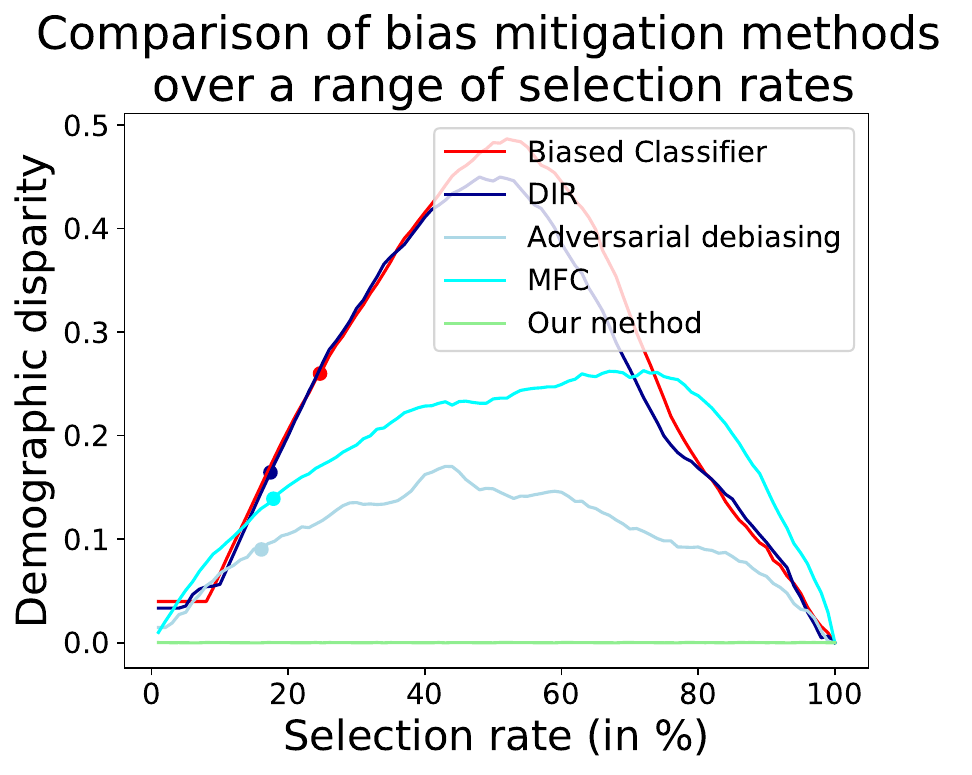}
    \caption[]{
An illustration of how demographic parity varies with selection rate. We apply a range of classical bias mitigation methods from AIF360~\cite{bellamy2018ai} and observe how fairness varies with selection rate. The dots indicate default thresholds. As the thresholds vary, fairness oscillates wildly (results on the Adult dataset, details Section~\ref{subsec:bias_mit_methods}). \emph{Just as standard fairness methods fail to consider the selection rate, existing analyses of the cost of fairness~\cite{corbett2017algorithmic,friedler2019comparative, haas2019price, von2021cost} also fail to take it into account.}\label{fig:DP_SR_plot}} 
\end{figure}

However, limited resources are not taken into account in most discussions in fair machine learning. This is illustrated in Figure~\ref{fig:DP_SR_plot}, where for many bias mitigation methods fairness oscillates wildly over different resource levels.
\citet{kwegyir2023repairing} highlight how practitioners typically need to adapt the threshold to ensure that outcomes meet their domain-specific needs, while \citet{corbett2023measure} argue that the trade-offs in fair decision-making are most acute when there are constrained resources. 
The inability to account for real-world constraints 
may contribute to the tiny amount of fair algorithms publically being deployed. 
This contrasts starkly with the abundance of studies that use fairness metrics solely for the evaluation of deployed systems \cite{buolamwini2018gender, barocas2023fairness}.

In the typical setting of an unconstrained budget, \citet{mittelstadt2023unfairness} posit that increasing harm to advantaged groups solely to achieve fairness is not an optimal approach and refer to this as `leveling down'. We agree with this viewpoint, but argue that within contexts constrained by limited resources, rebalancing is often necessary to redistribute the resources. In these circumstances, deliberately not fully utilizing resources would be a form of `leveling down'.
In this study, we make the following practical and theoretical contributions:
\begin{itemize}
    \item We formulate fair machine learning as a resource-constrained problem, where we treat positive decisions as a resource to allocate among different groups. 
    \item We provide a  connection to \textit{leveling up} \cite{mittelstadt2023unfairness}, showing that under constrained resources, leveling up results in a solution that enforces equality in harm between groups. 
    \item We extend prior work on the cost of fairness \citep{corbett2017algorithmic} by quantifying this cost within resource-constrained settings. Using underlying parameters, we derive mathematical bounds and empirically validate these patterns.  
\end{itemize}

\section{Background} \label{sec:background}
We consider fairness in classification. Starting with a classifier \(c_w(\cdot)\) parameterized by weights \(w\), we
let \harm~ be a measure of expected harm of a
classifier \(c_w(\cdot)\) over a particular distribution \(\mathcal D\) where $y_x$ represents the true target label of instance $x$,
i.e., 
\begin{equation}
  \harm = \mathrm{E}_{x\in \mathcal D} H(y_x,c_w(x))  
\end{equation}
Such measures of a harm \harm~might be $ 1-\text{precision}$ (for example, when measuring the proportion of people incorrectly stopped by the police); or $1 - \text{recall}$ (when measuring the number of cancer cases unflagged for follow-up treatment). 

 In fair classification, we typically measure fairness with respect to a protected attribute, such as gender or ethnicity. Using this protected attribute, we can partition the dataset into groups.
Group fairness metrics measure the (in)equality between these groups with respect to some measure of harm. For example, equal opportunity requires the recall between groups to be equal, while demographic parity enforces an equal selection rate~\cite{verma2018fairness}. Typically, a fair classifier is found by minimizing some global loss $\ell$ (such as accuracy or a continuous proxy such as the logistic loss) while ensuring that the harm is the same per group.

This means that we are searching for a solution to the following problem (where $\mathcal G$ is a partitioning of the distribution into groups with respect to a particular protected attribute):
\begin{equation}
\begin{aligned} \label{eq:fair_class}
    &\min_w \, \mathrm{E}_{x\in \mathcal D} \ell(y_x,c_w(x))  \\
    \text{ such that }
    &\harm[g_1]=\harm[g_2] \, \forall g_1,g_2\in \mathcal G 
\end{aligned}
\end{equation}
Notions of fairness include demographic parity, where the harm corresponds to $1-$ the selection rate, and equal opportunity where it corresponds to the false negative rate.

\subsection{Leveling Down}
\citet{mittelstadt2023unfairness} observe that methods to enforce fairness often level down; that is, they may enforce fairness by decreasing harm in some groups, but also by increasing harm to other groups (e.g.~naively enforcing Equal Opportunity while detecting cancer will often result in a higher false negative rate for some groups -- they receive a lower rate of cancer detection than they would otherwise). This process of enforcing fairness can alter the overall selection rate of the model~\cite{goethals2024reranking}, and is distinct from any leveling down that might be inherently required by the resource constraints discussed earlier.
To this end, they suggested replacing Equation~\ref{eq:fair_class} with rate constraints, which enforced that the harm in question (e.g. being denied follow-up care) should be below a certain level $h$ for every group:
\begin{align} \label{eq_harmbelowk} 
    \min_w \, \mathrm{E}_{x\in \mathcal D} \ell(y_x,c_w(x)) \\ \text{ such that } \harm[g] \leq h\, \forall g \in \mathcal G
\end{align}
This is the same as saying that in the worst case, \emph{the harm should be below $h$, for any group}.

The challenge now becomes ``How should $h$ be set?'' When deploying a particular model, stakeholders and data scientists often have to agree on acceptable global levels of harm, for example, an acceptable recall rate (referred to as sensitivity in the medical literature) for early cancer detection. A similar process can also select a maximal per group harm.

This notion of leveling-up, and decreasing the harm in Equation~\ref{eq_harmbelowk} is related to but distinct from minimax-fair machine learning, which minimizes the error for the worst-off group~\cite{martinez2020minimax,diana2021minimaxgroupfairnessalgorithms,abernethy2020active}. In both minimax fairness and leveling-up the concern is with reducing the harm to the worst-off group and only increasing harms to other groups when necessary, but in minimax fairness the harm is exclusively assumed to be caused by a lack of accuracy or high log-loss (this leads to a formulation of minimize the maximal loss, hence minimax). In comparison, leveling-up is concerned with decreasing other harms such as per-group recall or selection rate, which are distinct from accuracy. \footnote{Both notions are closely aligned to the philosophical principle of the maximin rule~\cite{rawls2017theory}. According to Rawls, resources should be distributed in a way that maximizes the benefits for the least-advantaged members of society. However, as a fundamental limitation, neither minimax nor leveling-up considers what should be done under constrained resources.}

Taking a step back, we can ask the questions: ``\textit{If $H$ is a harm, why not set $h$ to zero? If particular groups are being harmed by not being selected, why not select everyone and not bother with machine learning? Similarly, if people are being harmed by a failure to detect cancer, why not schedule everyone for follow-up testing?}''

There are multiple possible answers here. One is a matter of personal utility that harms are generally not one-sided. Failure to repay a loan can lead to bankruptcy for both the lender and customer, resulting in devastating personal consequences. Scheduling unnecessary medical tests is at best alarming, and in the worst case can result in death, depending on how intrusive the follow-up tests are. These types of considerations have been addressed by prior work on the cost of fairness, where personal utility is measured by accuracy~\cite{menon2018cost}, and \citet{bakalar2021fairness}, which measures the effect of the actual implementation of fairness principles in real-world situations. 

Constrained resources provide another answer. 
Ideally, we would prefer to offer scholarships to every student and fast-track treatment for every patient. However, the real world often lacks the resources needed to achieve this.
Understanding how fairness choices are limited and guided by resource constraints is thus crucial to move fair machine learning from the theoretical to a production environment.

Fair classification under limited resources is related to the domain of fair allocation within welfare economics, which focuses on ensuring that resources such as time or physical goods are distributed among actors in a way that meets certain criteria~\cite{bertsimas2011price,bertsimas2012efficiency, donahue2020fairness, rambachan2020economic,sinclair2022sequential,banerjee2022fair}. Another related area is fair ranking, which often aims to ensure fairness with respect to a top-k selection~\cite{zehlike2017fa}.

\section{Constrained Resources} \label{subsec:constrained_resources}
We formalize the notion of \emph{resources}, as the proportion of instances predicted as positive by a machine learning model. This term can be used interchangeably with \emph{capacity}, or with \emph{selection rate}, \emph{positive decision rate}, or \emph{positive prediction rate} when talking about the proportion of positively predicted instances.

\begin{proposition}
Under constrained resources, maximally leveling-up results in equality of harm between groups.
\end{proposition}

This means that when we are optimizing the distribution of a specified number of resources and are minimizing the harm experienced by any group (leveling up), this would lead to equality of harm between groups.

\begin{proof}
We define \rate~as the selection rate (i.e.~the expected proportion of positive decisions) of a classifier over a distribution \(\mathcal D\). 
We consider the problem of optimizing the distribution of limited resources, denoted by $r$, in order to minimize the maximum harm experienced by any group. This optimization problem is formulated as:
    \begin{equation}
    \label{eq:oxonfair}
        \min \max_g \harm[g] \quad \text{ such that } \quad \rate \leq r
    \end{equation}
where \rate~is a weighted sum of group-specific selection rates \rate[g] with all weights positive, thus making \rate~strictly increasing with respect to each \rate[g]. 

\paragraph*{Case 1: Harm is strictly decreasing with respect to $\rate[g]$}
We first consider the case where \harm[g] is strictly decreasing as a function of \rate[g] (for example, if the harm is $1 - \text{recall}$, we expect the harm to be strictly decreasing when more members of the group are selected).  Since \harm[g] can be written as an invertible function of \rate[g], it follows that \rate[g] is also strictly decreasing with respect to \harm[g].

Given that \harm[g] is strictly decreasing with \rate[g], and $r_D$ is strictly increasing with respect to \rate[g], it follows that \harm[g] is strictly decreasing with respect to $r_D$.

We define $W$ as the set of worst-off groups, i.e., the groups experiencing the maximum harm, such that:
\begin{equation}
    W = \arg\max_g \harm[g]
\end{equation}
The optimum must occur when \( \rate = r \). If not, one could distribute the remaining resources such that the rates \rate[g] for the worst-off groups increase without exceeding r, reducing the maximum harm (since \harm[g] is decreasing with respect to \rate[g]). Therefore the optimal solution uses all the available resources.

Furthermore, the optimum is reached when the harm levels are equal across all groups (so all groups are in $W$). If this were not the case, then there exists at least one group that is not in $W$ and that could sacrifice some of its resources to the worse-off groups. Adjusting the allocation in such a way that the group not in $W$ receives less, while groups at maximal harm receive more, would lead to a reduced overall maximum harm due to the strict decrease of \harm[g] with \rate[g].
Thus, under these conditions, equality in harm distribution is enforced in the optimal solution, ensuring that the distribution of resources maximally benefits the least advantaged groups within the constraints set by \( \rate \leq r \).

\paragraph*{Case 2: Harm is strictly increasing with respect to $\rate[g]$}
Now consider the case where harm is strictly increasing with respect to $\rate[g]$.  An example of this might be a harm function like $1 - \text{precision}$, where increasing the selection rate for group $g$ leads to higher harm.
In this case, the optimal solution is trivial with an overall selection rate of 0, as this would minimize harm across all groups. Equality is (trivially) achieved as no group receives any resources. 
However, in this case, it might be more realistic to flip the sign and require the overall selection rate to be at least some r. We can simply replace the classifier $c_w$ with its negation, and now find that the harm is strictly decreasing with respect to the selection rate of the new classifier. Revisiting the previous proof, we find that equality holds.

Hence, equality must hold in all four cases (strictly increasing or strictly decreasing harm, and a global selection rate that is either constrained above or below some value).
\end{proof}

\subsection{Cost of Fairness}
Before we proceed further, we  note that assigning a cost to actions is inherently a political action, and that often these costs reflect the beliefs of data scientists and other professionals as much as they do the raw data.

Indeed, one justification for constraints such as demographic parity is that ground-truth data is often gender- or ethnically-biased, and in some circumstances we can \emph{a priori} expect uniform rates across these populations (in line with the ``We're All Equal'' world-view of \cite{friedler2021possibility}, which asserts that there are no innate differences between groups). 
As such, any estimate of cost comes with the usual caveat of
\emph{assuming the ground-truth is correct.} But even as simplified
approximations, these costs remain useful for understanding the potential
trade-offs in enforcing fairness.
With this in mind, we consider the following question:
\emph{What is the global change in harm from an optimal classifier when we minimise the harm of the worst-off group?}

Under constrained resources, this has the same result as enforcing equality of harms between groups. We can measure this by the difference between the harm of the optimal classifier $c^o_w(.)$ and the classifier that satisfies fairness $c^f_w(.)$.
\begin{equation*}
\harmfn[c^f_w]-\harmfn[c^o_w]
\end{equation*}
 While other studies have previously analyzed the cost of fairness~\cite{corbett2017algorithmic,friedler2019comparative, haas2019price, von2021cost}, they did not consider the implicit trade-off that comes from constrained resources in decision-making, but instead measured how classifiers deteriorate with  fairness. Some of the calculated costs may very well arise from selecting a different \textit{number} of people; however, we will keep this number constant and focus instead on the costs associated with selecting a different \textit{set} of individuals.

We might also be interested in performance metrics outside the fairness metric enforced.
Of particular interest is the change in precision that occurs if we minimize the harm to the worst-off group.
This represents a common scenario. In a medical context, it corresponds to the
question: \emph{What proportion of healthy patients instead of sick will
we see as we increase test sensitivity for disadvantaged groups?}

This leads to a follow-up question:
\emph{What is the increase or decrease in selection rate needed to preserve the current rate of global harm, if we enforce equality?}
The last question represents a common political solution to this problem.
When particular groups are disadvantaged by the status quo, it is often easier to increase the resources, and target them at the disadvantaged groups, rather than requiring currently advantaged groups to accept less access to resources.

\section{Bounding the cost of fairness}
\label{sec:methods}

\subsection{Bounding the Cost of Fairness} \label{subsec:cost_bounds}
We now formalize these costs and consider the cost of the change induced by enforcing fairness at a particular selection rate.
As we are modeling the cost of inducing fairness under rate constraints, we consider the changes in the labeling induced by swaps, where the label assigned to one instance changes from positive to negative, and the label assigned to another instance changes from negative to positive. A sequence of swaps must always preserve the global selection rate, and any pair of labellings with the same selection rate, can be transformed from one to another by a sequence of swaps.
We write $p$ for the proportion of instances in the dataset swapped from positive to negative (or equivalently the proportion swapped from negative to positive), and $c$ for the average cost of swapping a pair.

By definition, the cost of inducing fairness is:
\begin{equation}
    \text{cost}= p \cdot c
\end{equation}
Note that $p\in[0,0.5]$, and for standard linear measures (e.g. accuracy, recall, specificity) 
the second term is bounded by
\begin{equation}
c=\frac{\text{\#instances in dataset}}
        {\text{\#instances used in the measure}}
\end{equation}
i.e. $c=1$ for accuracy, and, writing $b$ for the label base rate, $c=b^{-1}$ for Recall and False Positive Rate, and $c=(1-b)^{-1}$ for Specificity and False Negative Rate. As such, $pc$ and $c/2$ are both upper bounds on the cost of fairness. 

For example, at a given global selection rate $r$, the cost of fairness, in terms of change in accuracy, is at most the proportion of swaps ($p$); while the cost in terms of change of recall is at most  $p/b$.

We can create successive bounds based on this identity:
Given $r$, $p$ is bounded above by $r$ (all positive instances are swapped); and $(1-r)$ (all negative instances are swapped). Therefore, the cost of fairness is bounded above by $rc$, and $(1-r)c$. As such, if $r$ or $(1-r)$ is much smaller than $c^{-1}$, there is little cost in enforcing fairness. 

This matches our empirical finding in Section~\ref{sec:enf_fairness} that the cost is typically bell-shaped with respect to the global selection-rate. The highest costs occur when the selection rates are closest to the selection rates of an unconstrained classifier. This means that many of the existing works on the cost of fairness~\cite{menon2018cost,friedler2019comparative, hort2023bias} overestimate the cost of fairness for scenarios with other resource levels. 

For scenarios where the selection rate substantially exceeds the base rate, this is because for any informative classifier, the majority of candidates that might be positive would be selected early on, regardless of which group they belong to, and the remaining pool of candidates are likely to have negative labels regardless of which group they belong to. 
These bounds are much weaker for non-linear measures such as precision, if they exist. For precision, assuming a fixed global selection rate $r$, we have $c=r^-1$. Consequentially, the low selection-rate bound given by $rc=1$ is vacuous. However, other bounds remain informative when there is a higher selection rate -- particularly, the bounds $(1-r)c^{-1}$ and $gc^{-1}$, indicate that we should see similar patterns in the costs with respect to precision for higher selection rates.  Empirically, we find this --- precision exhibits similar trends to the linear measures of recall and accuracy. Reflecting the much weaker bound for low selection rates, it sometimes exhibits strong changes at low resource levels, which are not present for accuracy and recall. This can be seen in Figures~\ref{fig:compas_loss} and \ref{fig:law_loss}.

Similarly, given $g$, the proportion of the dataset occupied by the smallest group, we know $p<g$, and the cost of inducing fairness is bounded above by $gc$.\footnote{Relabelling only a subset of the smallest group is not always sufficient to enforce fairness, e.g. consider equal precision, but it holds for linear measures.} Again, if $g$ is much smaller than $c$, there is little cost in inducing fairness. This finding is supported in our experiments (see Figures~\ref{fig:ss_dp} and \ref{fig:ss_eo}).

Finally, looking at the second term $c$ (the average cost of relabeling), we find that this second factor explains much of what we will see in Figure~\ref{fig:parameter_exp}. For example, increasing \emph{global} noise decreases performance for both groups, at broadly similar proportions. This overall decrease shrinks the gap between the per-group performance, reducing the average cost of relabelling points.  Similarly, increasing classification noise for the group with greatest classification error, or increasing the difference in base rates, increases the performance gap between groups and consequentially the cost of relabelling. 

Following this analysis, it is reasonable to ask if bounds exist for \emph{levelling up}, and if there are bounded increases in selection rate for a disadvantaged group guaranteed to remove unfairness (while keeping the selection rate of the advantaged group as-is). In brief, the answer is yes for demographic parity; for equal opportunity you may need to relabel all datapoints in a disadvantaged group; and levelling up need not be possible for equal precision. \footnote{These results differ from Section~\ref{subsec:cost_bounds} because the previous analysis was advantaged by the stability of measures -- we altered per group selection rates and hoped that global measures remained stable. In this new case, we are trying to alter per-group measures and the stability of measures is now a disadvantage.}

For demographic parity, there are no stability concerns. Let $g$ be the proportion of the dataset corresponding to the disadvantaged group, and $r$ the global classifier selection rate. Then an upper bound for the selection rate excluding members of the disadvantaged group is $r/(1-g)$, and at most we require an additional $gr/(1-g)$ proportion of points to be positively labelled to obtain parity.
For equal opportunity, only instances assigned a positive ground-truth label alter the recall, and it is possible that a badly-trained classifier selects all negatively labeled  instances first, and only then positively labelled points. If we consider a classifier like this, and a group with only one positively labelled point (in the ground truth), all points within the group must be selected to get any non-zero recall rate for that group. As such, $g$ is a bound.
For equal precision, we consider the same example, and note that while you must select all datapoints to obtain non-zero precision, the precision will be the base rate of the group, and it may be insufficient for equality with respect to the other group.

\subsection{Varying Global Selection Rates} \label{sec:discussion}

Until now, we assumed the resources to be fixed, but practitioners may find it feasible to increase or decrease them, particularly if the resource levels were initially determined under an unfair model allocation. For instance, in healthcare, additional investments could be directed towards expanding screening facilities to counterbalance the decrease in true positives, effectively maintaining the detection rates prior to implementing fairness measures. 
Similarly, in educational settings such as student admission, institutions might adjust the size of admitted cohorts, either decreasing it to preserve the average quality, or increasing it to have the same number of graduates. 
\cite{donahue2020fairness} discuss how increasing the level of available resources is a critical goal where advocacy and political action can play a key role. 

\section{Enforcing Fairness} \label{sec:enf_fairness}
Fairness at different selection rates can be enforced using two existing approaches. \citet{kwegyir2023repairing} enforces equal opportunity and demographic parity by aligning per-group score distributions while minimizing the earthmovers difference. The fairness toolkit of \citet{delaney2024oxonfair} optimizes an objective subject to a constraint (such as maximize min group recall, subject to the overall selection rate being less than 0.7). 
All methods fall into the family of post-hoc bias mitigation strategies, that attempt to make the output of machine learning model fair after the model has been trained~\cite{mehrabi2021survey}. 
For completeness, we set out our strategy for efficiently finding thresholds, however, this has limited novelty and should not be considered a significant contribution. 

To measure the cost of fairness, we implement a per-group parameter sweep, based on the harms-based analysis of Section~\ref{sec:background}.
Given a classifier $c_w$, we say it makes a positive decision regarding datapoint $x$ if $c_w(x)>t$, where $t$ is some threshold. We consider fairness as in Equation~\ref{eq:fair_class} where some notion of harm \harm~ is being equalized across groups.
Now, we simply vary a per group threshold $t_g$, while computing the group harm \harm[g], and the corresponding selection rate \rate[g]. We consider some measure of acceptable harm $h$, and select the set of per group thresholds $t_g$ so that the harm per group lies just under this $h$. This is (approximately) fair with respect to harm $H$, and has a global selection rate
\begin{equation}
r=\frac{\sum_{g} |g|\cdot \ratet[g]}{\sum_{g} |g|}    
\end{equation}
where $|g|$ is the number of datapoints in group $g$.
By sweeping over possible values $h$, and examining the corresponding global selection rate, we find a fair solution that lies just under a target selection rate. The corresponding thresholds can be used to enforce fairness on previously unseen test data. We conduct an analysis across a range of selection rates $r$ (from 1\% to 100 \%) representing different resource levels. This process is particularly straightforward for demographic parity. In this case, the harm corresponds to 1- the selection rate, and the process simplifies to just selecting the top $r$ from each group.\footnote{For demographic parity, ideal thresholds can be calculated on the test set, but for most other measures the thresholds need to be calculated on a separate validation set, as they require access to the target label in order to compute the harms. Owing to sampling error, thresholds selected on the validation data do not correspond perfectly to selection rates on unseen data. As such, to ensure that the number of instances predicted as positive exactly matches the available resources, we calculate the proportion of resources to allocate to each group for each resource level $R$ on the validation set, and transfer these proportions to the test set.}

\section{The Empirical Cost of Fairness} \label{sec:results}
We discuss the datasets, classifiers and metrics in Appendix~\ref{section:mm}.~\footnote{The code for the experiments is publicly available at \url{https://github.com/SofieGoethals/RCF}.} In Table~\ref{tab:model_perf}, we analyze the AUC performance of a machine learning model for different groups: the entire population, and separately for disadvantaged and advantaged groups. Additionally, we report the average cost of enforcing fairness as the loss in precision between the default allocation and the fair allocation, averaged over all selection rates.\footnote{The results for recall and accuracy are in line with precision and can be found in Table~\ref{tab:model_perf_recacc}}
We see that the average cost of enforcing fairness is the lowest for the Fitzpatrick17K dataset, which also has a very low base rate disparity (see Table~\ref{tab:datasets}). The average cost is also very low for the Law dataset, despite its significant base rate disparity. A possible reason for this could be the relatively small size of the disadvantaged group in this dataset, which results in less reallocation of resources.
The average cost of enforcing fairness across other datasets is more similar, however it is worth noting that the cost associated with enforcing both fairness metrics on CelebA is higher than for the other datasets. 
It is hard to attribute the difference in costs between the datasets to a single factor, as many of the parameters will be different. This is why we perform a separate analysis on the Adult Income dataset.

\begin{table}[htb]
\centering
\caption{Model performance (entire population, advantaged group and disadvantaged group) and the average cost of fairness (loss in precision) when enforcing DP and EO. The average is calculated over all the selection rates from $1\%$ to $100\%$.}
\label{tab:model_perf}
 \begin{adjustbox}{max width=\linewidth}
\begin{tabular}{c|cccccc}
Dataset& Adult& Compas& Dutch&Law & CelebA & Fitz17K \\ \hline
AUC&   0.896& 0.812& 0.917&0.870 & 0.957 & 0.826\\
$\text{AUC}_{priv}$& 
0.887 & 0.797& 0.884&0.847 & 0.931 & 0.829\\
$\text{AUC}_{prot}$& 0.871 & 0.793& 0.914 &0.858 & 0.969 & 0.814\\
Avg. cost~(DP) & 0.025 & 0.026 & 0.028 & 0.005 & 0.069 & 0.001 \\
 Avg. cost~(EO) & 0.009 & 0.014 & 0.007 & 0.003 & 0.009 & 0.000\\
 \end{tabular}
 \end{adjustbox}
\end{table}

\begin{figure*}
        \begin{subfigure}[htb]{0.24\textwidth}
            \includegraphics[width=\textwidth]{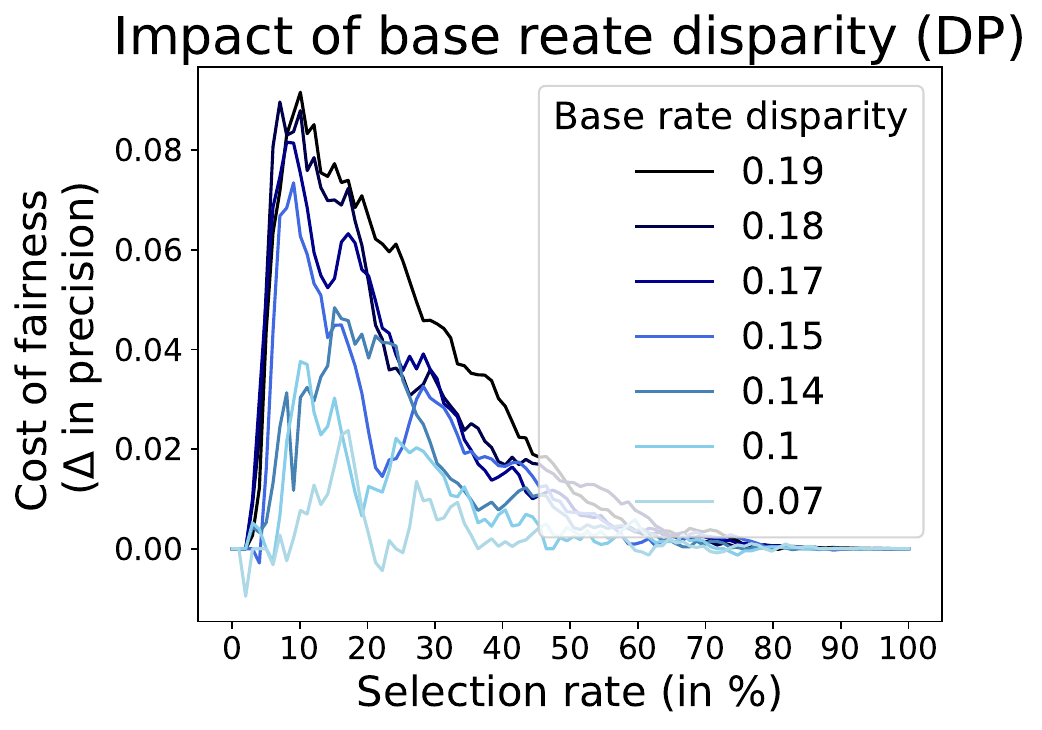}
            \caption{Base rate disparity (DP)}
            \label{fig:br_dp}
        \end{subfigure}
        \hfill 
        \begin{subfigure}[htb]{0.24\textwidth}
            \includegraphics[width=\textwidth]{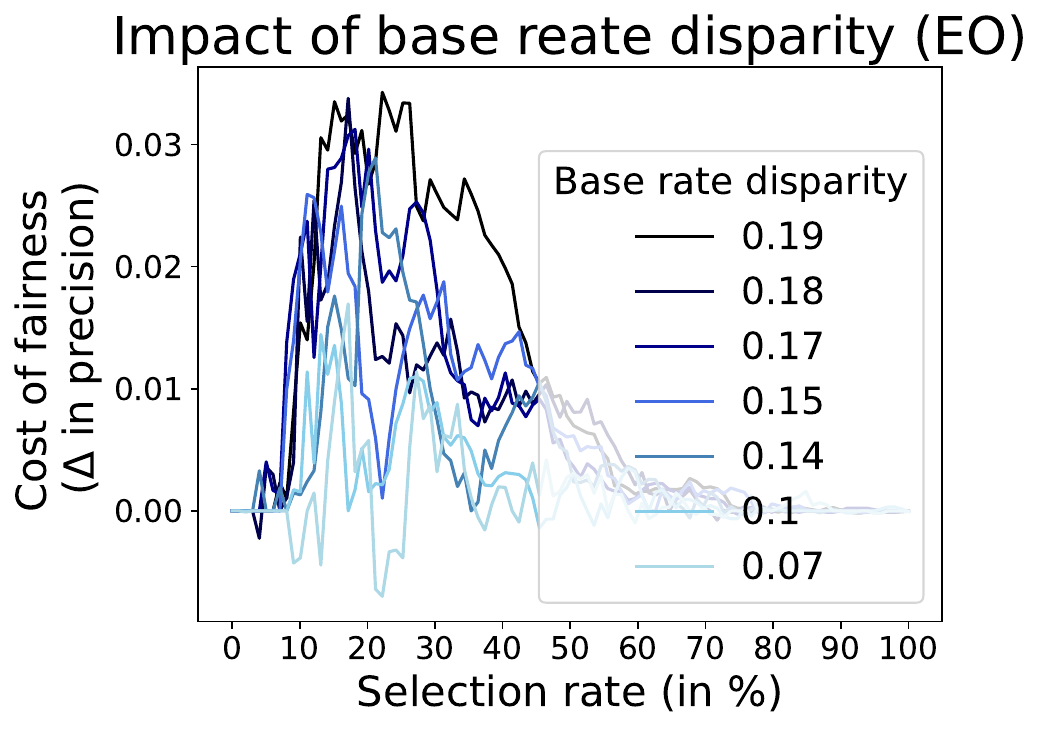}
            \caption{Base rate disparity (EO)}
            \label{fig:br_eo}
        \end{subfigure}
        \begin{subfigure}[htb]{0.24\textwidth}
            \includegraphics[width=\textwidth]{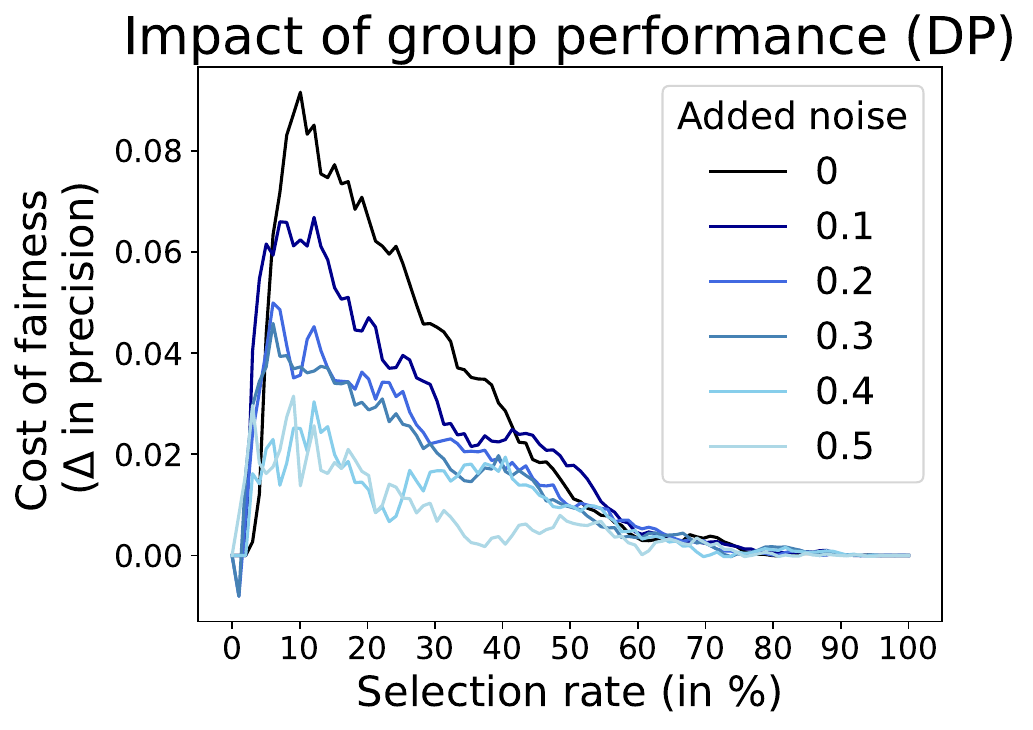}
            \caption{Noise (DP)}
            \label{fig:gn_dp}
        \end{subfigure}
        \hfill 
        \begin{subfigure}[htb]{0.24\textwidth}
            \includegraphics[width=\textwidth]{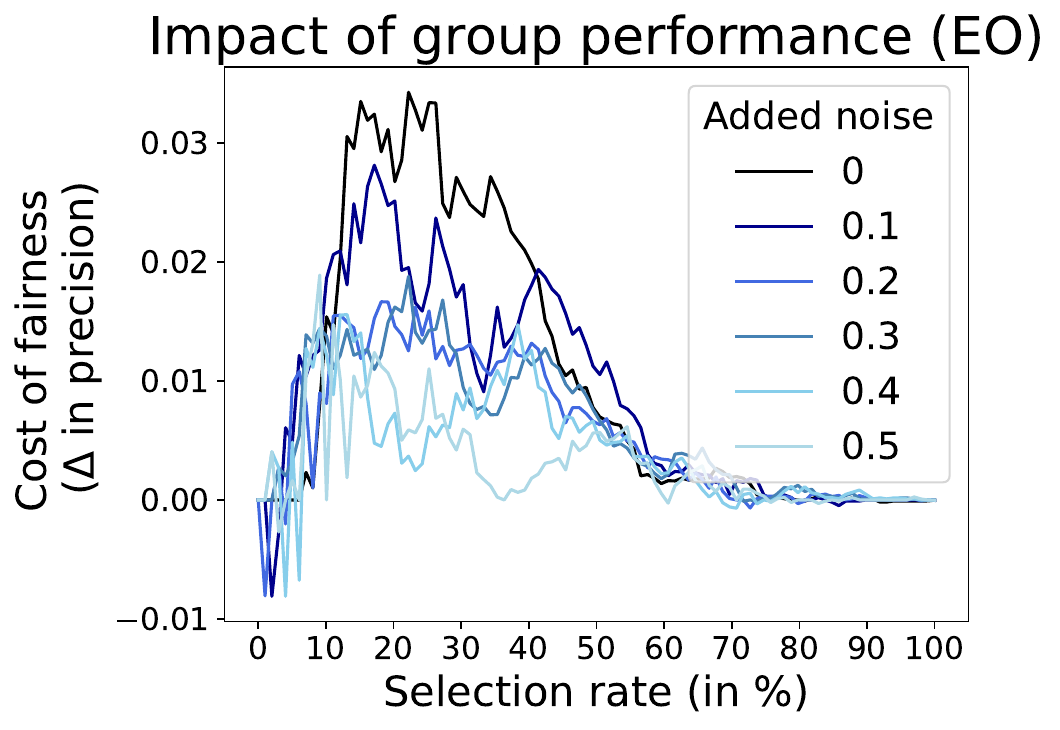}
            \caption{Noise (EO)}
            \label{fig:gn_eo}
        \end{subfigure}
                \begin{subfigure}[htb]{0.24\textwidth}
            \includegraphics[width=\textwidth]{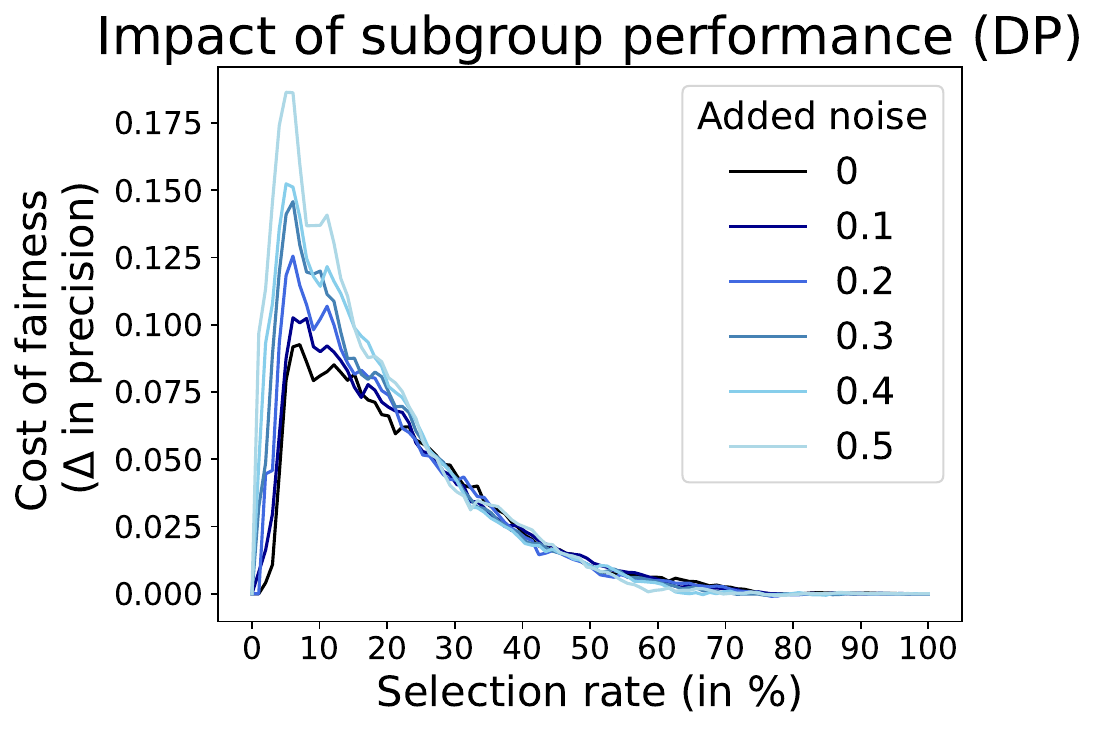}
            \caption{Subgroup noise (DP)}
            \label{fig:sn_dp}
        \end{subfigure}
        \hfill 
        \begin{subfigure}[htb]{0.24\textwidth}
            \includegraphics[width=\textwidth]{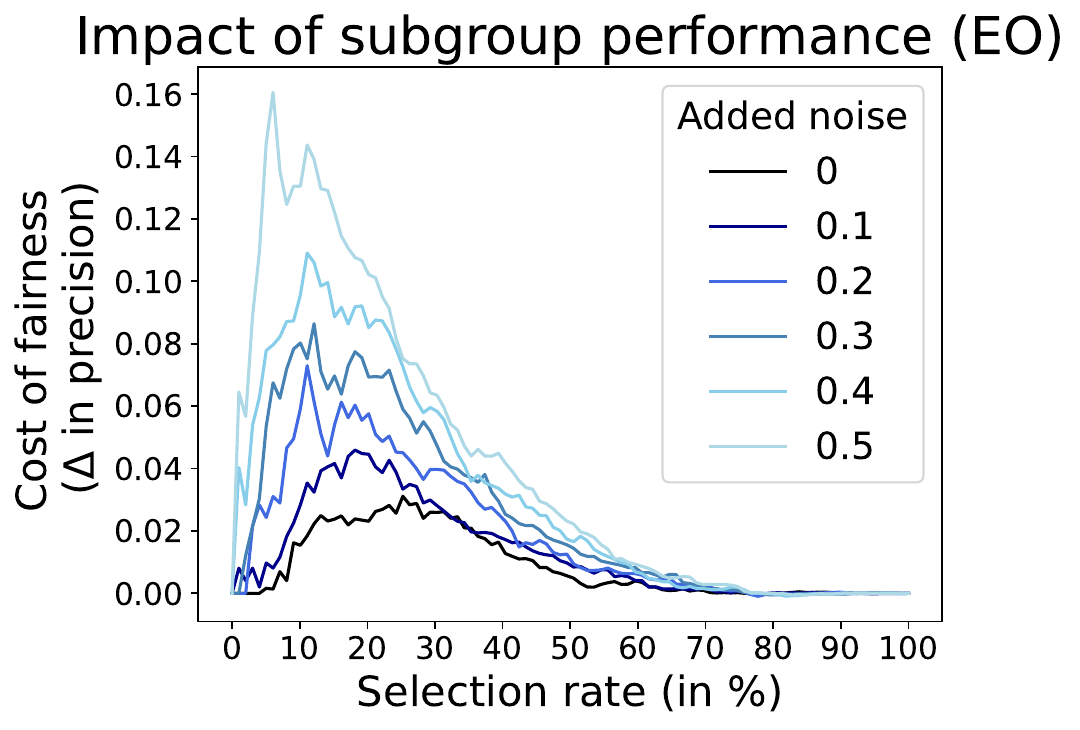}
            \caption{Subgroup noise (EO)}
            \label{fig:sn_eo}
        \end{subfigure}
        \begin{subfigure}[htb]{0.24\textwidth}
        \includegraphics[width=\textwidth]{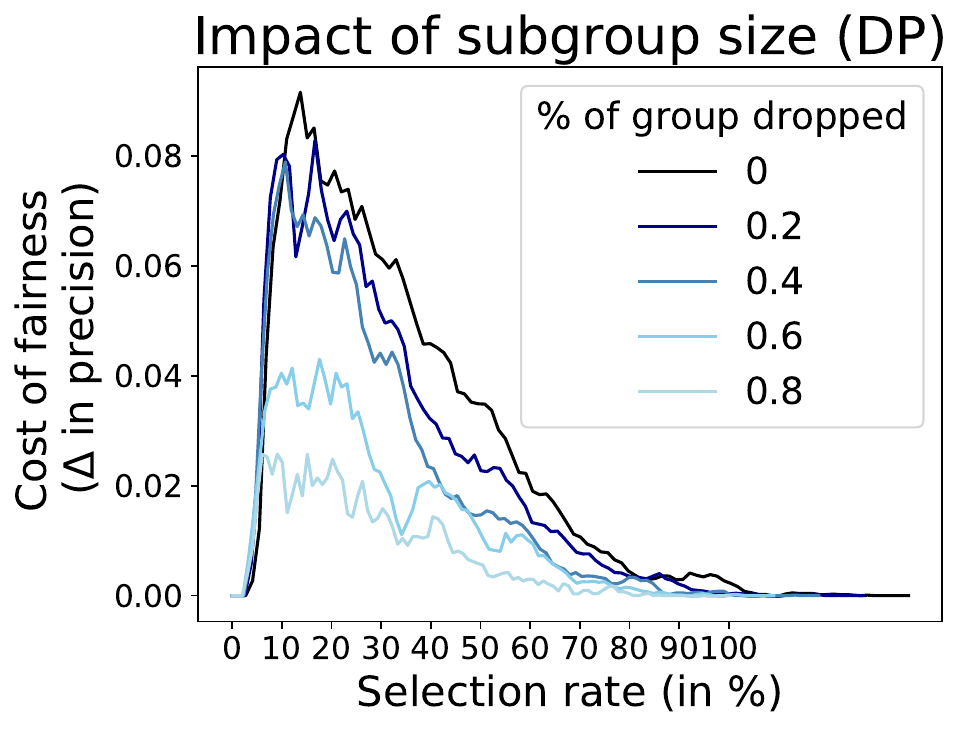}
        \caption{Subgroup size (DP)}
        \label{fig:ss_dp}
    \end{subfigure}
    \hfill 
    \begin{subfigure}[htb]{0.24\textwidth}
        \includegraphics[width=\textwidth]{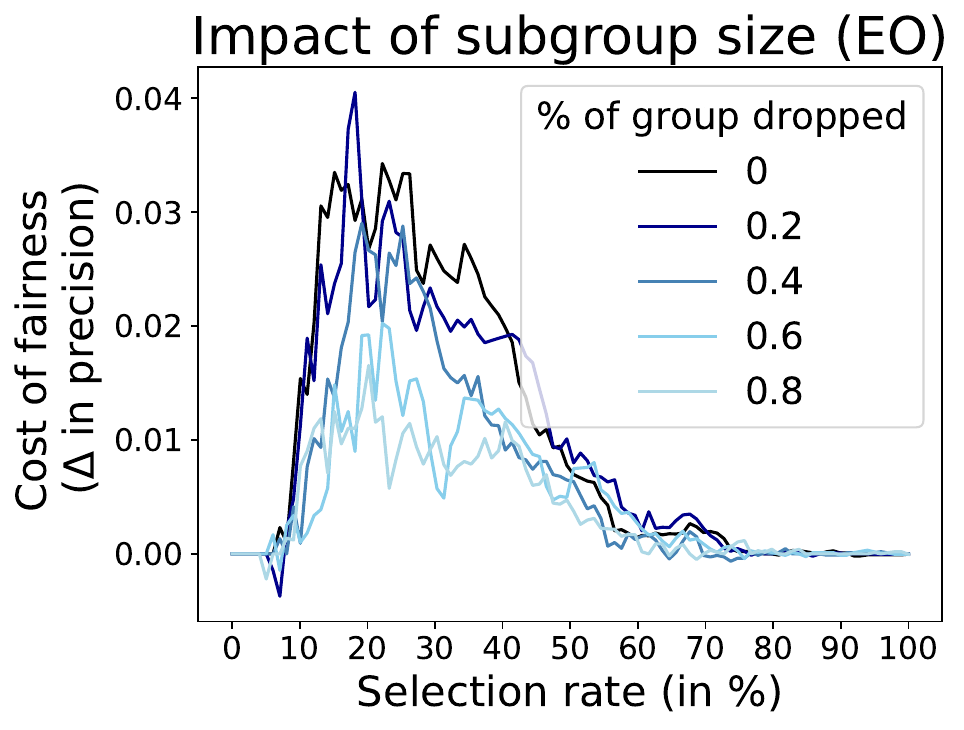}
        \caption{Subgroup size (EO)}
        \label{fig:ss_eo}
    \end{subfigure}
    \caption{Impact of parameters on the cost of fairness, measured as the \textbf{loss in precision} (Adult Income dataset). We see that reducing the base rate disparity leads to a lower  cost of fairness. Similarly, introducing more noise across all groups, also lowers the cost of fairness, but introducing more noise exclusively in the disadvantaged group, increases the cost of fairness.
     Finally, reducing the size of the disadvantaged group, results in a decrease in the cost of fairness as well.}
    \label{fig:parameter_exp}
\end{figure*}

Here, we systematically vary key parameters—one at a time—to observe their effects on the cost of fairness. Specifically, we explore modifications to the base rate disparity, the size of the disadvantaged group, the noisiness of the whole dataset and the noisiness of the disadvantaged group. The impacts of these changes are shown in Figure~\ref{fig:parameter_exp}, with the original dataset's results represented by a black line.

\textbf{Base rate disparity}
First, we alter the base rate disparity. We see in Figures~\ref{fig:br_dp} and \ref{fig:br_eo}, that a reduction in the base rate disparity leads to a lower cost of fairness. This trend is more pronounced when enforcing DP compared to EO, but it is evident under both fairness metrics.
\cite{menon2018cost} also find that the trade-off between accuracy and fairness is determined by the strength of the correlation between the sensitive attribute and the target.

\textbf{Noise}
We alter the performance of the model by introducing random noise (with varying degrees from 0 to 0.5) to the feature values of the whole dataset. The base rates of both groups remain the same, but the model struggles more with correctly classifying individuals,  resulting in a lower AUC score. 
We see that as model performance decreases, the average cost of fairness also goes down, both for DP and EO (Figures~\ref{fig:gn_dp}-\ref{fig:gn_eo}).
This global decrease shrinks the gap between the performances of each group, reducing the overall cost of fairness.
This phenomenon might also explain why the average cost of fairness is so high for the CelebA dataset in Table~\ref{tab:model_perf}. The model is very good in distinguishing negatives and positives from each other (as measured by the AUC), so the average cost will be high.\footnote{Following this reasoning, we see that if we would fit a perfect model (so the model predicts the target label perfectly for all instances) to a dataset with some level of base rate disparity, then the maximum cost of enforcing demographic parity would be reached at a point between the base rates and be exactly equal to the level of base rate disparity in that dataset. In the case of CelebA, the model is not perfect but the maximum cost is also relatively close to the level of the base rate disparity, and is reached at a point between the base rates.}

\textbf{Subgroup noise}
We can also add noise only to the members of the disadvantaged group. This lowers the performance of the model for the disadvantaged group, but not for members of the advantaged group.
As shown in Figures~\ref{fig:sn_dp} and \ref{fig:sn_eo}, this modification increases the cost of fairness for both DP and EO, with a larger effect for EO. 
\cite{chen2018my} also find that when there is a difference in noise level, and available covariates are not equally predictive of the outcome in both groups, fairness cannot be satisfied without sacrificing accuracy.
\cite{dutta2020there} confirm that if there is not enough separability information for one group compared to the other, being fair will reduce the accuracy.

\textbf{Size of the disadvantaged group}
Lastly, the size of the disadvantaged group substantially influences the cost of fairness. When the disadvantaged group is smaller, fewer adjustments are necessary to achieve fairness, thereby reducing the cost.
This relationship is clearly illustrated in Figures~\ref{fig:ss_dp} and \ref{fig:ss_eo}, where reducing the size of the disadvantaged group, by randomly excluding part of this group, consistently lowers the cost of fairness.
This is an expected finding and aligns with our bounds analysis in Section~\ref{subsec:cost_bounds}, however, we have not seen it discussed previously.

\subsection{The Impact of Selection Rate}

\begin{figure*}[htb]
    \centering
    \begin{subfigure}[b]{0.31\textwidth}
        \centering
        \includegraphics[width=\textwidth]{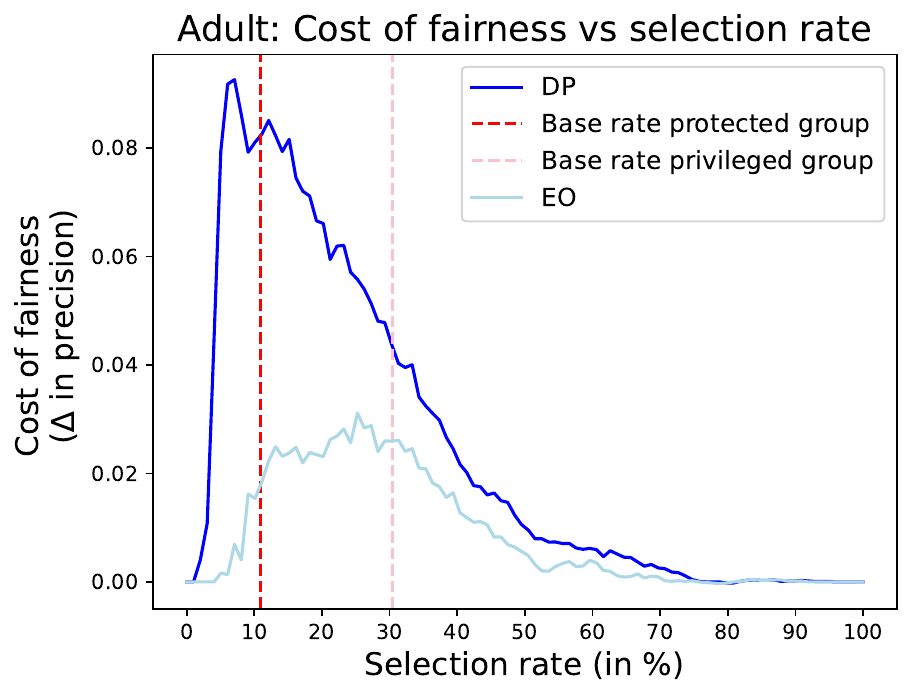}
        \caption{Adult}
        \label{fig:adult_loss}
    \end{subfigure}
    \hfill
    \begin{subfigure}[b]{0.31\textwidth}
        \centering
        \includegraphics[width=\textwidth]{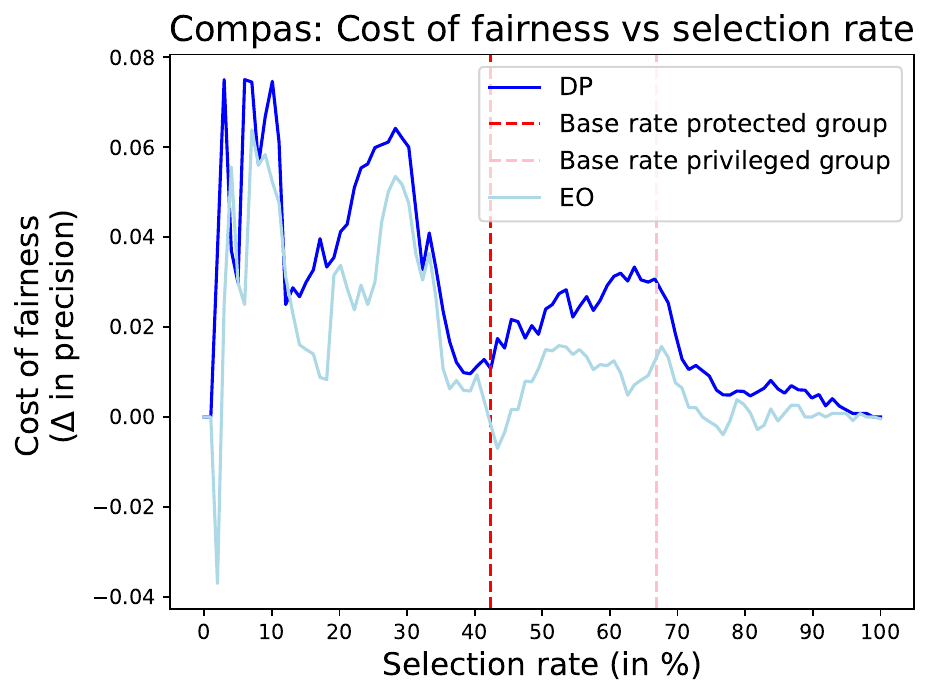}
        \caption{Compas}
        \label{fig:compas_loss}
    \end{subfigure}
    \hfill
    \begin{subfigure}[b]{0.31\textwidth}
        \centering
        \includegraphics[width=\textwidth]{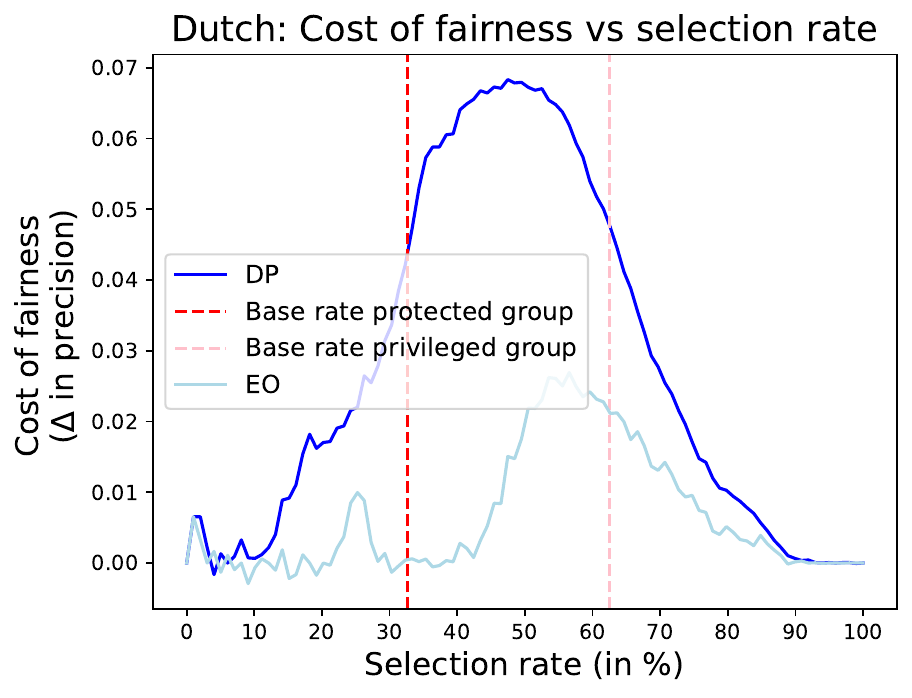}
        \caption{Dutch}
        \label{fig:dutch_loss}
    \end{subfigure}
    
    \vspace{1em} 
    
    \begin{subfigure}[b]{0.31\textwidth}
        \centering
        \includegraphics[width=\textwidth]{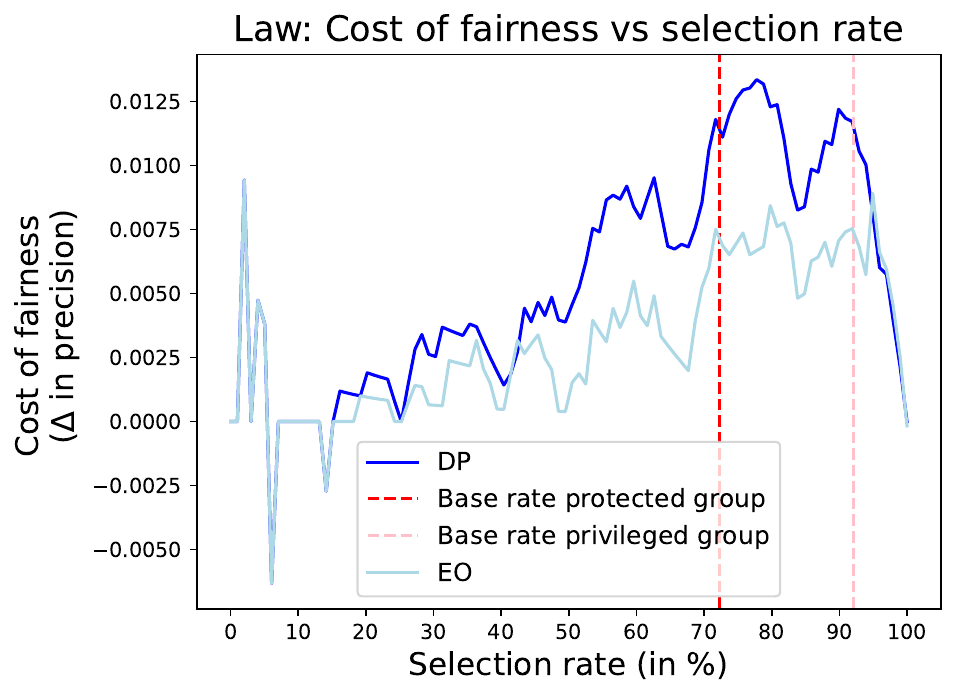}
        \caption{Law}
        \label{fig:law_loss}
    \end{subfigure}
    \hfill
    \begin{subfigure}[b]{0.31\textwidth}
        \centering
        \includegraphics[width=\textwidth]{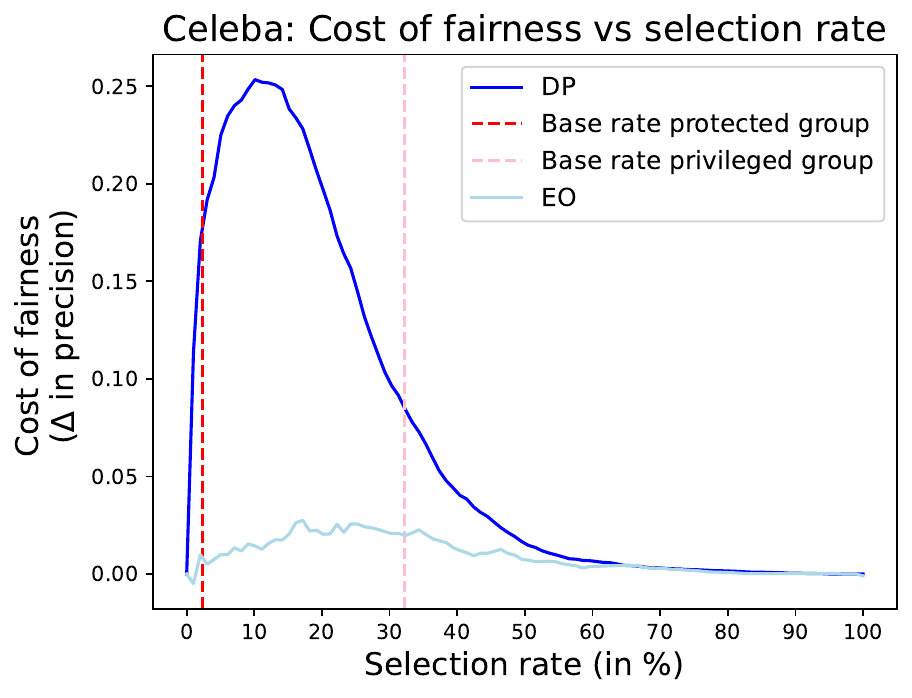}
        \caption{CelebA - Earrings}
        \label{fig:celeba_loss}
    \end{subfigure}
    \hfill
    \begin{subfigure}[b]{0.31\textwidth}
        \centering
        \includegraphics[width=\textwidth]{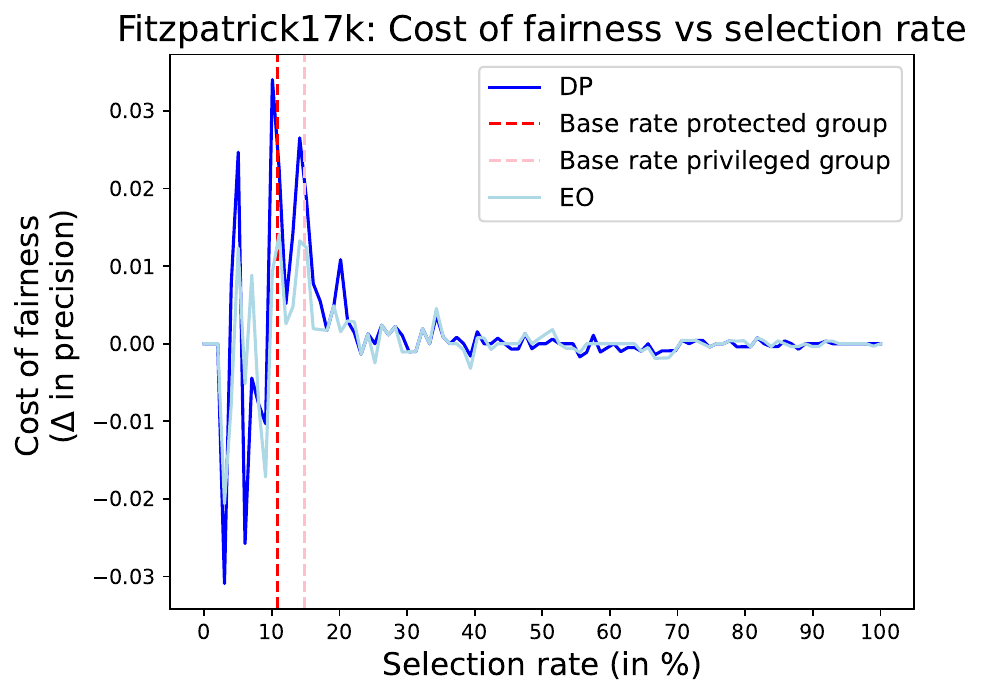}
        \caption{Fitzpatrick17k}
        \label{fig:fitzpatrick_loss}
    \end{subfigure}
    
    \caption{Cost of fairness (loss in \textbf{precision}) for all datasets when enforcing demographic parity (DP) and equal opportunity (EO). The cost of fairness for both metrics heavily depends on the available resource level and thus the used selection rate.}
    \label{fig:datasets_loss_precision}
\end{figure*}

Figures~\ref{fig:adult_loss}-\ref{fig:fitzpatrick_loss} demonstrate the loss in precision for various selection rates when using a fair allocation compared to the default allocation of the machine learning model. We see that this cost varies substantially depending on the available resources.
For the Adult (Figure~\ref{fig:adult_loss}), Compas (Figure~\ref{fig:compas_loss}), CelebA (Figure~\ref{fig:celeba_loss}) and Fitzpatrick 17K (Figure~\ref{fig:fitzpatrick_loss}) datasets, the
cost is highest for low selection rates, while for the Dutch (Figure~\ref{fig:dutch_loss}) dataset, the cost is the highest for medium selection rates, and for the Law (Figure~\ref{fig:law_loss}) dataset, the cost is the highest for high selection rates. 
The influence of resource level on the cost of fairness has not been discussed before, yet it is a critical factor.
Depending on the resource level, the cost can either be negligible or very high. 
The results for recall and accuracy are in line with precision and can be found in Figures~\ref{fig:datasets_loss_recall} and \ref{fig:datasets_loss_accuracy}. 
We can also partially agree with \cite{rodolfa2021empirical} who posited that for very constrained top-$k$ settings, so when the selection rate is very low, the cost of fairness can be negligible. However, we only find this to be true when the base rate is a lot higher and the performance of the model is good enough. 
We get more insights into how these trade-offs behave for each resource level by analyzing the different scenarios in Section~\ref{sec:resource_levels}.

\subsection{The Impact of the Fairness Metric}

Table~\ref{tab:model_perf} and Figures~\ref{fig:datasets_loss_precision}–\ref{fig:adult_resourcelevels} show that the cost of fairness is generally lower when enforcing equal opportunity (EO) than demographic parity (DP) across all selection rates. This is because EO typically requires fewer instances to be ``swapped." \citet{liu2018delayed} also find that the selection rate enforced by EO is closer to the optimal, while \cite{hardt2016equality} show that enforcing EO leads to a lower loss in profits than enforcing DP.
However, EO can be costlier when ensuring an equal true positive rate imposes stricter constraints than an equal positive rate, for example when one group has lower data quality. In such cases, DP is less affected since it disregards prediction accuracy, whereas EO requires equally poor recall across groups, degrading overall performance. As shown in Figures~\ref{fig:sn_dp}- \ref{fig:sn_eo}, the cost of enforcing EO rises significantly faster than the cost of enforcing DP when one subgroup becomes noisier.

\section{Conclusion}
This work addresses the practical concern of enforcing fairness with a limited budget or, equivalently, as the selection rate (or threshold) varies.
We unified leveling up \cite{mittelstadt2023unfairness} and fairness under budgetary constraints, and have shown that the solutions for each approach coincide under constrained resources (Section \ref{subsec:constrained_resources}). We have also shown how simple methods for setting per group thresholds can enforce fairness under these rate constraints (Section \ref{sec:methods}).
Using this, we empirically measured the cost of fairness under rate constraints -- in terms of decreased performance, or as an increase in global harm; and derived theoretical bounds for these costs (Section \ref{sec:results}). While the bounds are not intended to be tight, they are informative, and align with our empirical findings.
This allows us to investigate the actual cost of fairness, rather than changes in performance metrics that are heavily influenced by a change in the overall selection rate~\cite{goethals2024reranking}. 
Compared to other fairness works, our analysis more closely aligns with real-world challenges, where resources are often fixed, and allows organizations to have reasonable expectations about what costs to expect and why.
Furthermore, we demonstrated that the decision-making context matters, particularly the behavior of the classifier; the used fairness metric; and the level of available resources. 

\section*{Acknowledgments}
This work has been supported through research funding provided by the Wellcome Trust (Grant number 223765/Z/21/Z), Sloan Foundation (Grant number G-2021-16779), Department of Health and Social Care, EPSRC (Grant number EP/Y019393/1), and Luminate Group. Their funding supports the Trustworthiness Auditing for AI project and Governance of Emerging Technologies research programme at the Oxford Internet Institute, University of Oxford.
Sofie Goethals was funded by Flemish Research Foundation (Grant number 11N7723N), and by a travel grant (Grant number V413024N) awarded by Flemish Research Foundation to visit the University of Oxford.

\bibliographystyle{abbrvnat}  
\bibliography{references}

\begin{thebibliography}{47}
\providecommand{\natexlab}[1]{#1}
\providecommand{\url}[1]{\texttt{#1}}
\expandafter\ifx\csname urlstyle\endcsname\relax
  \providecommand{\doi}[1]{doi: #1}\else
  \providecommand{\doi}{doi: \begingroup \urlstyle{rm}\Url}\fi

\bibitem[Abernethy et~al.(2020)Abernethy, Awasthi, Kleindessner, Morgenstern, Russell, and Zhang]{abernethy2020active}
J.~Abernethy, P.~Awasthi, M.~Kleindessner, J.~Morgenstern, C.~Russell, and J.~Zhang.
\newblock Active sampling for min-max fairness.
\newblock \emph{arXiv preprint arXiv:2006.06879}, 2020.

\bibitem[Bakalar et~al.(2021)Bakalar, Barreto, Bergman, Bogen, Chern, Corbett-Davies, Hall, Kloumann, Lam, Candela, et~al.]{bakalar2021fairness}
C.~Bakalar, R.~Barreto, S.~Bergman, M.~Bogen, B.~Chern, S.~Corbett-Davies, M.~Hall, I.~Kloumann, M.~Lam, J.~Q. Candela, et~al.
\newblock Fairness on the ground: Applying algorithmic fairness approaches to production systems.
\newblock \emph{arXiv preprint arXiv:2103.06172}, 2021.

\bibitem[Banerjee et~al.(2022)Banerjee, Eichhorn, and Kempe]{banerjee2022fair}
S.~Banerjee, M.~Eichhorn, and D.~Kempe.
\newblock Fair and efficient allocation with quotas.
\newblock \emph{arXiv preprint arXiv:2204.13019}, 2022.

\bibitem[Barocas et~al.(2023)Barocas, Hardt, and Narayanan]{barocas2023fairness}
S.~Barocas, M.~Hardt, and A.~Narayanan.
\newblock \emph{Fairness and machine learning: Limitations and opportunities}.
\newblock MIT press, 2023.

\bibitem[Bellamy et~al.(2018)Bellamy, Dey, Hind, Hoffman, Houde, Kannan, Lohia, Martino, Mehta, Mojsilovic, et~al.]{bellamy2018ai}
R.~K. Bellamy, K.~Dey, M.~Hind, S.~C. Hoffman, S.~Houde, K.~Kannan, P.~Lohia, J.~Martino, S.~Mehta, A.~Mojsilovic, et~al.
\newblock Ai fairness 360: An extensible toolkit for detecting, understanding, and mitigating unwanted algorithmic bias.
\newblock \emph{arXiv preprint arXiv:1810.01943}, 2018.

\bibitem[Bertsimas et~al.(2011)Bertsimas, Farias, and Trichakis]{bertsimas2011price}
D.~Bertsimas, V.~F. Farias, and N.~Trichakis.
\newblock The price of fairness.
\newblock \emph{Operations research}, 59\penalty0 (1):\penalty0 17--31, 2011.

\bibitem[Bertsimas et~al.(2012)Bertsimas, Farias, and Trichakis]{bertsimas2012efficiency}
D.~Bertsimas, V.~F. Farias, and N.~Trichakis.
\newblock On the efficiency-fairness trade-off.
\newblock \emph{Management Science}, 58\penalty0 (12):\penalty0 2234--2250, 2012.

\bibitem[Buolamwini and Gebru(2018)]{buolamwini2018gender}
J.~Buolamwini and T.~Gebru.
\newblock Gender shades: Intersectional accuracy disparities in commercial gender classification.
\newblock In \emph{Conference on fairness, accountability and transparency}, pages 77--91. PMLR, 2018.

\bibitem[Celis et~al.(2019)Celis, Huang, Keswani, and Vishnoi]{celis2019classification}
L.~E. Celis, L.~Huang, V.~Keswani, and N.~K. Vishnoi.
\newblock Classification with fairness constraints: A meta-algorithm with provable guarantees.
\newblock In \emph{Proceedings of the conference on fairness, accountability, and transparency}, pages 319--328, 2019.

\bibitem[Chen et~al.(2018)Chen, Johansson, and Sontag]{chen2018my}
I.~Chen, F.~D. Johansson, and D.~Sontag.
\newblock Why is my classifier discriminatory?
\newblock \emph{Advances in neural information processing systems}, 31, 2018.

\bibitem[Chen and Guestrin(2016)]{chen2016xgboost}
T.~Chen and C.~Guestrin.
\newblock Xgboost: A scalable tree boosting system.
\newblock In \emph{Proceedings of the 22nd acm sigkdd international conference on knowledge discovery and data mining}, pages 785--794, 2016.

\bibitem[Corbett-Davies et~al.(2017)Corbett-Davies, Pierson, Feller, Goel, and Huq]{corbett2017algorithmic}
S.~Corbett-Davies, E.~Pierson, A.~Feller, S.~Goel, and A.~Huq.
\newblock Algorithmic decision making and the cost of fairness.
\newblock In \emph{Proceedings of the 23rd acm sigkdd international conference on knowledge discovery and data mining}, pages 797--806, 2017.

\bibitem[Corbett-Davies et~al.(2023)Corbett-Davies, Gaebler, Nilforoshan, Shroff, and Goel]{corbett2023measure}
S.~Corbett-Davies, J.~D. Gaebler, H.~Nilforoshan, R.~Shroff, and S.~Goel.
\newblock The measure and mismeasure of fairness.
\newblock \emph{The Journal of Machine Learning Research}, 24\penalty0 (1):\penalty0 14730--14846, 2023.

\bibitem[Delaney et~al.(2024)Delaney, Fu, Wachter, Mittelstadt, and Russell]{delaney2024oxonfair}
E.~Delaney, Z.~Fu, S.~Wachter, B.~Mittelstadt, and C.~Russell.
\newblock Oxonfair: A flexible toolkit for algorithmic fairness.
\newblock \emph{arXiv preprint arXiv:2407.13710}, 2024.

\bibitem[Deng et~al.(2009)Deng, Dong, Socher, Li, Li, and Fei-Fei]{deng2009imagenet}
J.~Deng, W.~Dong, R.~Socher, L.-J. Li, K.~Li, and L.~Fei-Fei.
\newblock Imagenet: A large-scale hierarchical image database.
\newblock In \emph{2009 IEEE conference on computer vision and pattern recognition}, pages 248--255. Ieee, 2009.

\bibitem[Diana et~al.(2021)Diana, Gill, Kearns, Kenthapadi, and Roth]{diana2021minimaxgroupfairnessalgorithms}
E.~Diana, W.~Gill, M.~Kearns, K.~Kenthapadi, and A.~Roth.
\newblock Minimax group fairness: Algorithms and experiments, 2021.
\newblock URL \url{https://arxiv.org/abs/2011.03108}.

\bibitem[Donahue and Kleinberg(2020)]{donahue2020fairness}
K.~Donahue and J.~Kleinberg.
\newblock Fairness and utilization in allocating resources with uncertain demand.
\newblock In \emph{Proceedings of the 2020 conference on fairness, accountability, and transparency}, pages 658--668, 2020.

\bibitem[Dutta et~al.(2020)Dutta, Wei, Yueksel, Chen, Liu, and Varshney]{dutta2020there}
S.~Dutta, D.~Wei, H.~Yueksel, P.-Y. Chen, S.~Liu, and K.~Varshney.
\newblock Is there a trade-off between fairness and accuracy? a perspective using mismatched hypothesis testing.
\newblock In \emph{International conference on machine learning}, pages 2803--2813. PMLR, 2020.

\bibitem[Dwork et~al.(2012)Dwork, Hardt, Pitassi, Reingold, and Zemel]{dwork2012fairness}
C.~Dwork, M.~Hardt, T.~Pitassi, O.~Reingold, and R.~Zemel.
\newblock Fairness through awareness.
\newblock In \emph{Proceedings of the 3rd innovations in theoretical computer science conference}, pages 214--226, 2012.

\bibitem[Feldman et~al.(2015)Feldman, Friedler, Moeller, Scheidegger, and Venkatasubramanian]{feldman2015certifying}
M.~Feldman, S.~A. Friedler, J.~Moeller, C.~Scheidegger, and S.~Venkatasubramanian.
\newblock Certifying and removing disparate impact.
\newblock In \emph{proceedings of the 21th ACM SIGKDD international conference on knowledge discovery and data mining}, pages 259--268, 2015.

\bibitem[Friedler et~al.(2019)Friedler, Scheidegger, Venkatasubramanian, Choudhary, Hamilton, and Roth]{friedler2019comparative}
S.~A. Friedler, C.~Scheidegger, S.~Venkatasubramanian, S.~Choudhary, E.~P. Hamilton, and D.~Roth.
\newblock A comparative study of fairness-enhancing interventions in machine learning.
\newblock In \emph{Proceedings of the conference on fairness, accountability, and transparency}, pages 329--338, 2019.

\bibitem[Friedler et~al.(2021)Friedler, Scheidegger, and Venkatasubramanian]{friedler2021possibility}
S.~A. Friedler, C.~Scheidegger, and S.~Venkatasubramanian.
\newblock The (im) possibility of fairness: Different value systems require different mechanisms for fair decision making.
\newblock \emph{Communications of the ACM}, 64\penalty0 (4):\penalty0 136--143, 2021.

\bibitem[Goethals and Calders(2024)]{goethals2024reranking}
S.~Goethals and T.~Calders.
\newblock Reranking individuals: the effect of fair classification within-groups.
\newblock \emph{arXiv}, pages 1--16, 2024.

\bibitem[Groh et~al.(2021)Groh, Harris, Soenksen, Lau, Han, Kim, Koochek, and Badri]{groh2021evaluating}
M.~Groh, C.~Harris, L.~Soenksen, F.~Lau, R.~Han, A.~Kim, A.~Koochek, and O.~Badri.
\newblock Evaluating deep neural networks trained on clinical images in dermatology with the fitzpatrick 17k dataset.
\newblock In \emph{Proceedings of the IEEE/CVF Conference on Computer Vision and Pattern Recognition}, pages 1820--1828, 2021.

\bibitem[Haas(2019)]{haas2019price}
C.~Haas.
\newblock The price of fairness-a framework to explore trade-offs in algorithmic fairness.
\newblock 2019.

\bibitem[Hardt et~al.(2016)Hardt, Price, and Srebro]{hardt2016equality}
M.~Hardt, E.~Price, and N.~Srebro.
\newblock Equality of opportunity in supervised learning.
\newblock \emph{Advances in neural information processing systems}, 29, 2016.

\bibitem[He et~al.(2016)He, Zhang, Ren, and Sun]{he2016deep}
K.~He, X.~Zhang, S.~Ren, and J.~Sun.
\newblock Deep residual learning for image recognition.
\newblock In \emph{Proceedings of the IEEE conference on computer vision and pattern recognition}, pages 770--778, 2016.

\bibitem[Hort et~al.(2023)Hort, Chen, Zhang, Harman, and Sarro]{hort2023bias}
M.~Hort, Z.~Chen, J.~M. Zhang, M.~Harman, and F.~Sarro.
\newblock Bias mitigation for machine learning classifiers: A comprehensive survey.
\newblock \emph{ACM Journal on Responsible Computing}, 2023.

\bibitem[Kwegyir-Aggrey et~al.(2023)Kwegyir-Aggrey, Dai, Cooper, Dickerson, Hines, and Venkatasubramanian]{kwegyir2023repairing}
K.~Kwegyir-Aggrey, J.~Dai, A.~F. Cooper, J.~Dickerson, K.~Hines, and S.~Venkatasubramanian.
\newblock Repairing regressors for fair binary classification at any decision threshold.
\newblock In \emph{NeurIPS 2023 Workshop Optimal Transport and Machine Learning}, 2023.

\bibitem[Le~Quy et~al.(2022)Le~Quy, Roy, Iosifidis, Zhang, and Ntoutsi]{le2022survey}
T.~Le~Quy, A.~Roy, V.~Iosifidis, W.~Zhang, and E.~Ntoutsi.
\newblock A survey on datasets for fairness-aware machine learning.
\newblock \emph{Wiley Interdisciplinary Reviews: Data Mining and Knowledge Discovery}, 12\penalty0 (3):\penalty0 e1452, 2022.

\bibitem[Liu et~al.(2018)Liu, Dean, Rolf, Simchowitz, and Hardt]{liu2018delayed}
L.~T. Liu, S.~Dean, E.~Rolf, M.~Simchowitz, and M.~Hardt.
\newblock Delayed impact of fair machine learning.
\newblock In \emph{International Conference on Machine Learning}, pages 3150--3158. PMLR, 2018.

\bibitem[Liu et~al.(2015)Liu, Luo, Wang, and Tang]{liu2015deep}
Z.~Liu, P.~Luo, X.~Wang, and X.~Tang.
\newblock Deep learning face attributes in the wild.
\newblock In \emph{Proceedings of the IEEE international conference on computer vision}, pages 3730--3738, 2015.

\bibitem[Martinez et~al.(2020)Martinez, Bertran, and Sapiro]{martinez2020minimax}
N.~Martinez, M.~Bertran, and G.~Sapiro.
\newblock Minimax pareto fairness: A multi objective perspective.
\newblock In \emph{International conference on machine learning}, pages 6755--6764. PMLR, 2020.

\bibitem[Mehrabi et~al.(2021)Mehrabi, Morstatter, Saxena, Lerman, and Galstyan]{mehrabi2021survey}
N.~Mehrabi, F.~Morstatter, N.~Saxena, K.~Lerman, and A.~Galstyan.
\newblock A survey on bias and fairness in machine learning.
\newblock \emph{ACM computing surveys (CSUR)}, 54\penalty0 (6):\penalty0 1--35, 2021.

\bibitem[Menon and Williamson(2018)]{menon2018cost}
A.~K. Menon and R.~C. Williamson.
\newblock The cost of fairness in binary classification.
\newblock In \emph{Conference on Fairness, accountability and transparency}, pages 107--118. PMLR, 2018.

\bibitem[Mittelstadt et~al.(2023)Mittelstadt, Wachter, and Russell]{mittelstadt2023unfairness}
B.~Mittelstadt, S.~Wachter, and C.~Russell.
\newblock The unfairness of fair machine learning: Levelling down and strict egalitarianism by default.
\newblock \emph{arXiv preprint arXiv:2302.02404}, 2023.

\bibitem[Rambachan et~al.(2020)Rambachan, Kleinberg, Mullainathan, and Ludwig]{rambachan2020economic}
A.~Rambachan, J.~Kleinberg, S.~Mullainathan, and J.~Ludwig.
\newblock An economic approach to regulating algorithms.
\newblock Technical report, National Bureau of Economic Research, 2020.

\bibitem[Rawls(2017)]{rawls2017theory}
J.~Rawls.
\newblock A theory of justice.
\newblock In \emph{Applied ethics}, pages 21--29. Routledge, 2017.

\bibitem[Rodolfa et~al.(2021)Rodolfa, Lamba, and Ghani]{rodolfa2021empirical}
K.~T. Rodolfa, H.~Lamba, and R.~Ghani.
\newblock Empirical observation of negligible fairness--accuracy trade-offs in machine learning for public policy.
\newblock \emph{Nature Machine Intelligence}, 3\penalty0 (10):\penalty0 896--904, 2021.

\bibitem[Sinclair et~al.(2022)Sinclair, Banerjee, and Yu]{sinclair2022sequential}
S.~R. Sinclair, S.~Banerjee, and C.~L. Yu.
\newblock Sequential fair allocation: Achieving the optimal envy-efficiency tradeoff curve.
\newblock \emph{ACM SIGMETRICS Performance Evaluation Review}, 50\penalty0 (1):\penalty0 95--96, 2022.

\bibitem[Verma and Rubin(2018)]{verma2018fairness}
S.~Verma and J.~Rubin.
\newblock Fairness definitions explained.
\newblock In \emph{Proceedings of the international workshop on software fairness}, pages 1--7, 2018.

\bibitem[von Zahn et~al.(2021)von Zahn, Feuerriegel, and Kuehl]{von2021cost}
M.~von Zahn, S.~Feuerriegel, and N.~Kuehl.
\newblock The cost of fairness in ai: Evidence from e-commerce.
\newblock \emph{Business \& information systems engineering}, pages 1--14, 2021.

\bibitem[Wachter et~al.(2020)Wachter, Mittelstadt, and Russell]{wachter2020bias}
S.~Wachter, B.~Mittelstadt, and C.~Russell.
\newblock Bias preservation in machine learning: the legality of fairness metrics under eu non-discrimination law.
\newblock \emph{W. Va. L. Rev.}, 123:\penalty0 735, 2020.

\bibitem[Wang et~al.(2020)Wang, Qinami, Karakozis, Genova, Nair, Hata, and Russakovsky]{wang2020towards}
Z.~Wang, K.~Qinami, I.~C. Karakozis, K.~Genova, P.~Nair, K.~Hata, and O.~Russakovsky.
\newblock Towards fairness in visual recognition: Effective strategies for bias mitigation.
\newblock In \emph{Proceedings of the IEEE/CVF conference on computer vision and pattern recognition}, pages 8919--8928, 2020.

\bibitem[Zehlike et~al.(2017)Zehlike, Bonchi, Castillo, Hajian, Megahed, and Baeza-Yates]{zehlike2017fa}
M.~Zehlike, F.~Bonchi, C.~Castillo, S.~Hajian, M.~Megahed, and R.~Baeza-Yates.
\newblock Fa* ir: A fair top-k ranking algorithm.
\newblock In \emph{Proceedings of the 2017 ACM on Conference on Information and Knowledge Management}, pages 1569--1578, 2017.

\bibitem[Zhang et~al.(2018)Zhang, Lemoine, and Mitchell]{zhang2018mitigating}
B.~H. Zhang, B.~Lemoine, and M.~Mitchell.
\newblock Mitigating unwanted biases with adversarial learning.
\newblock In \emph{Proceedings of the 2018 AAAI/ACM Conference on AI, Ethics, and Society}, pages 335--340, 2018.

\bibitem[Zong et~al.(2022)Zong, Yang, and Hospedales]{zong2022medfair}
Y.~Zong, Y.~Yang, and T.~Hospedales.
\newblock Medfair: benchmarking fairness for medical imaging.
\newblock \emph{arXiv preprint arXiv:2210.01725}, 2022.

\end{thebibliography}

\newpage
\appendix
\onecolumn

\section{Materials and Methods} \label{section:mm}
\subsection{Materials}

We use several real-world datasets common in tabular fair machine learning~\cite{le2022survey}. We also extend our findings to two datasets (CelebA \cite{liu2015deep} and Fitzpatrick17k \cite{groh2021evaluating}) from computer vision.
The \textbf{Adult Income} dataset is taken from from the 1994 census data, and contains a prediction task identifying if an individual's annual income surpasses \$50,000. The \textbf{Compas} dataset gathers demographic details and criminal records of defendants from Broward County to forecast the likelihood of reoffending within two years. The \textbf{Dutch Census} dataset from 2001 captures aggregated demographic data in the Netherlands, utilized to determine if an individual's occupation falls into a high-level (prestigious) or low-level category. The \textbf{Law Admission} dataset contains data from a 1991 survey by the Law School Admission Council (LSAC) across 163 U.S. law schools, aimed at predicting a student's success on the bar exam.
The \textbf{CelebA} dataset comprises over 200,000 celebrity images, each annotated with 40 attribute labels ranging from hair color to emotions, along with 5 landmark locations. We use the attribute `wearing earrings' as target label, as it is highly skewed towards the non Male class \cite{wang2020towards} and thus has a significant base rate disparity.
The \textbf{Fitzpatrick17k} dataset consists of approximately 17,000 dermatologist-curated skin lesion images, categorized across the Fitzpatrick skin type scale. We preprocess the data and binarize the labels following the approach of \cite{zong2022medfair}. 
We specify the protected and target attributes in Table~\ref{tab:datasets}.
We also report the \textbf{base rate disparity} for each dataset, which is defined as the actual difference in the proportion of positive outcomes between groups in a dataset. If we consider a binary sensitive attribute $S$, with $ns$ representing the disadvantaged group and $s$ the advantaged group it can be mathematically expressed as:
\begin{equation*}
    P(Y = 1 \mid S = ns) - P(Y = 1 \mid S = s)
\end{equation*}

\begin{table}[ht]
\onecolumn
        \caption{Used datasets. The values between brackets represent the percentage of the dataset with that value.}
    \label{tab:datasets}
     \centering
    \begin{adjustbox}{width=\linewidth}
    \begin{tabular}{l|llllll}
    \hline
         Name&  \# instances&  \# attributes &Protected attribute  &disadvantaged group&  Target attribute & Base rate disparity \\ \hline
         Adult&  48,842&  10 &Gender  &Female ($33.15\%)$&  High income (23.93\%) & 19.46\%\\ 
         Compas&  5,278&   7&Race  &African-American ($60.15\%$) &  Low risk (52.12\%) & 24.60\%\\
         Dutch Census&  60,420&   11&Gender  &Female ($50.10\%$)&  High occupation (47.60\%) & 29.86\%\\
         Law admission&  20,798&  11 &Race &Non-White ($15.90\%$)&  Pass the bar (88.98\%) & 19.82\%\\
         CelebA (image) & 202,599 & NA & Gender & Male (38.65\%) & Wear earrings (20.66\%) & 29.84\% \\
         Fitzpatrick17K (image) & 16,012 & NA & Skin color & Black (31.79\%) & Skin cancer (13.65\%) & 3.99\% \\
    \end{tabular}
    \end{adjustbox}
\end{table}

\subsection{Methods}
We use a standard train-test split,  where the model is trained on the training set and the predictive performance and cost of fairness is evaluated on a separate test set. 

For the tabular datasets, we use eXtreme Gradient Boosting (XGBoost) 
\cite{chen2016xgboost}. 
We optimize the number of boosting rounds through 5-fold cross-validation on the training set with early stopping after 10 rounds if no improvement is seen. The optimized model is trained on the entire training set and evaluates the probability of positive class outcomes. 
For the image datasets, we employ a Resnet-50 (CelebA) and Resnet-18 (Fitzpatrick-17k) backbone \cite{he2016deep} pretrained on ImageNet \cite{deng2009imagenet} for feature extraction. 

\subsubsection{Comparison with other bias mitigation methods} \label{subsec:bias_mit_methods}

Figure \ref{fig:DP_SR_plot} illustrates how demographic parity, one of the most common notions of fairness, varies with selection rate, when deploying a set of popular bias mitigation methods. We use Disparate Impact Remover (DIR)~\cite{feldman2015certifying}, Adversarial Debiasing (ADV)~\cite{zhang2018mitigating} and Meta Fair Classifier (MFC)~\cite{celis2019classification}, and use the implementation of AIF360~\cite{bellamy2018ai}.
The points represent the labels of the bias mitigation methods at their default threshold. However, as the thresholds vary, fairness oscillates wildly, revealing that these methods may break when considering the whole range of possible selection rates. For our experiments, we use a kind of threshold optimizer, which enforces fairness exactly at each possible selection rate.

\subsection{Metrics} \label{subsec:metrics}
\subsubsection{Performance metrics}

Each classifier returns a prediction score \( S \), where higher \( S \) values indicate a greater likelihood of \( Y = 1 \). This score is converted to a binary prediction using a threshold: \( \hat{Y} = \mathds{1}\{S > t\} \). Evaluating classifier performance with the prediction scores, unaffected by the threshold choice, is best measured by the Area Under the ROC Curve (AUC). The AUC score is given by:

\[
P(S(x_{i}) > S(x_{z}) \mid y_{i}=1, y_{z}=0)
\]

This formula measures the probability that the classifier ranks a randomly chosen positive instance higher than a randomly chosen negative instance.

We evaluate the performance of the default allocation and the fair allocation over a range of selection rates (where the number of selected instances is the same for both allocations). 
When the selection rate of a machine learning model is fixed, it implies that a set proportion of the instances will be chosen based on the highest scores predicted by the model, regardless of their actual scores.
A useful performance metric to measure is the proportion of relevant instances among the top $R$ instances selected by the model (where $R$ denotes the resource level). This means the number of selected instances that actually have a positive target label.
\begin{equation*}
    \text{Precision at $R$} = \frac{\text{Number of actual positives in the top $R$}}{R}
\end{equation*}

\begin{figure*}[htb]
    \centering
        \begin{subfigure}[b]{0.31\textwidth}
            \includegraphics[width=\textwidth]{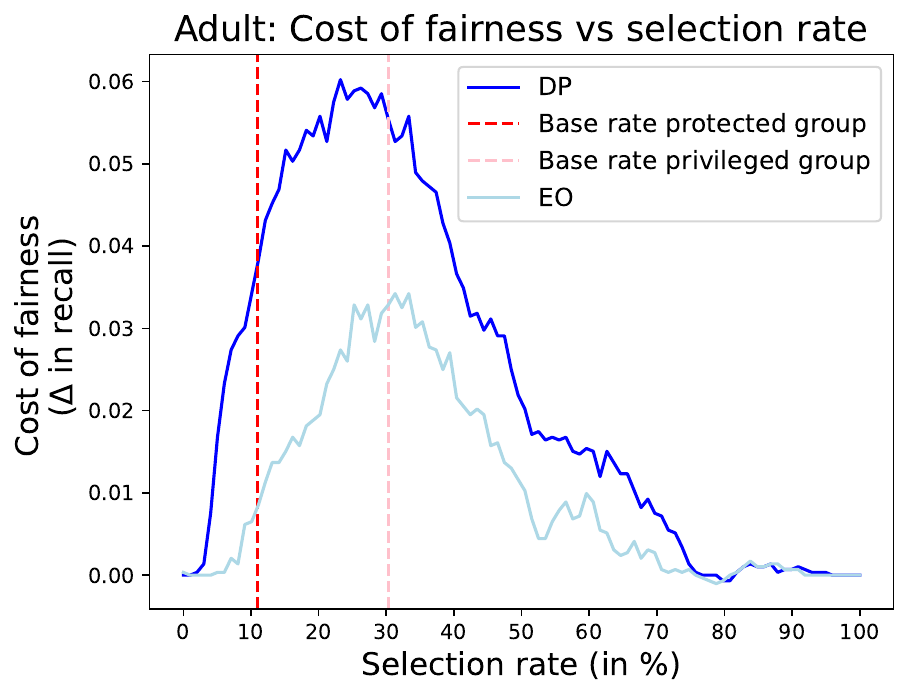}
            \caption{Adult}
        \end{subfigure}
        \begin{subfigure}[b]{0.31\textwidth}
            \includegraphics[width=\textwidth]{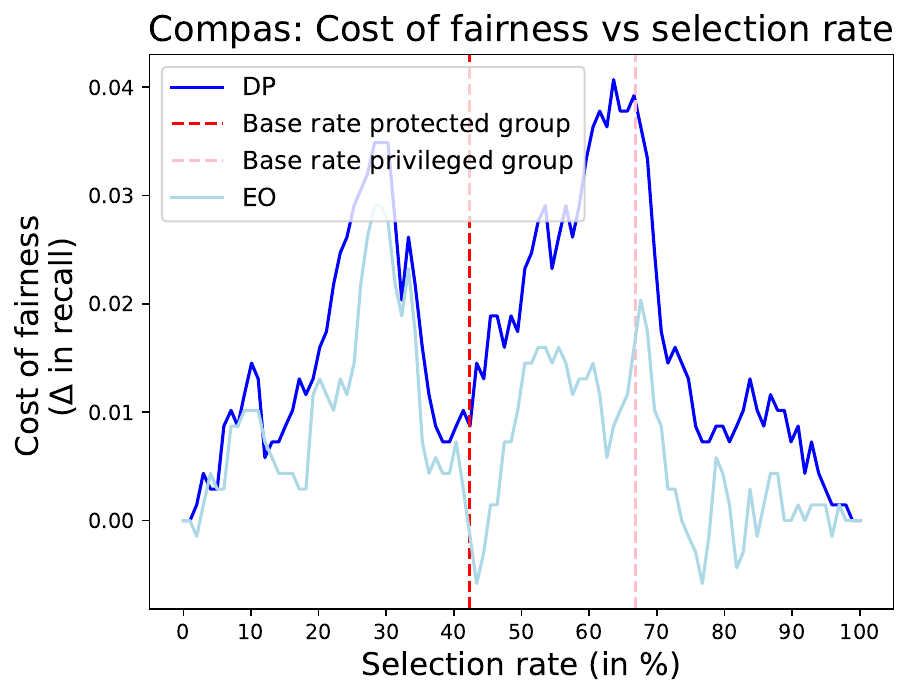}
            \caption{Compas}
        \end{subfigure}
        \begin{subfigure}[b]{0.31\textwidth}
            \includegraphics[width=\textwidth]{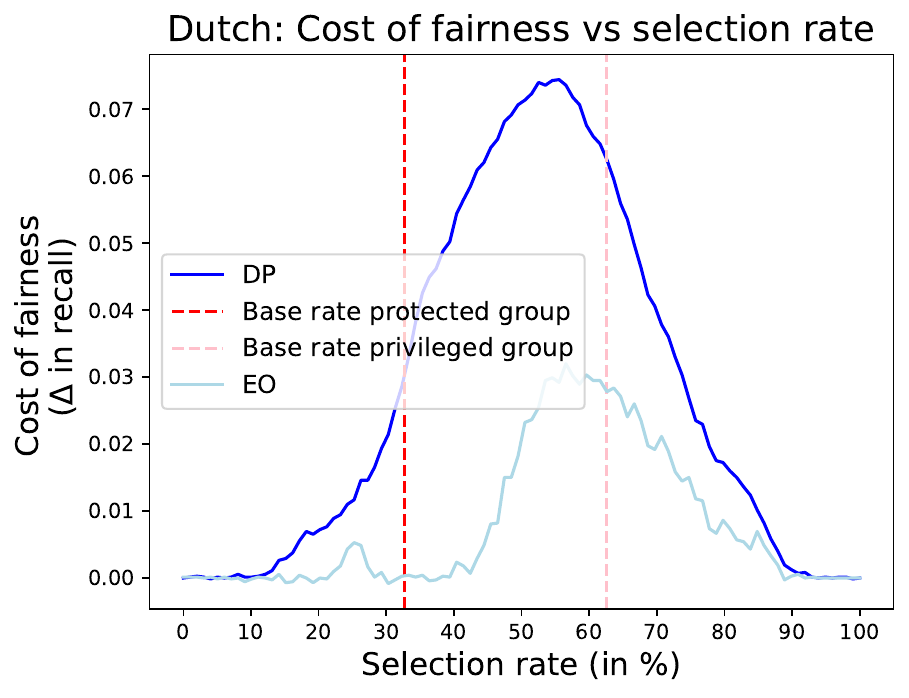}
            \caption{Dutch}
        \end{subfigure}
        \begin{subfigure}[b]{0.31\textwidth}
            \includegraphics[width=\textwidth]{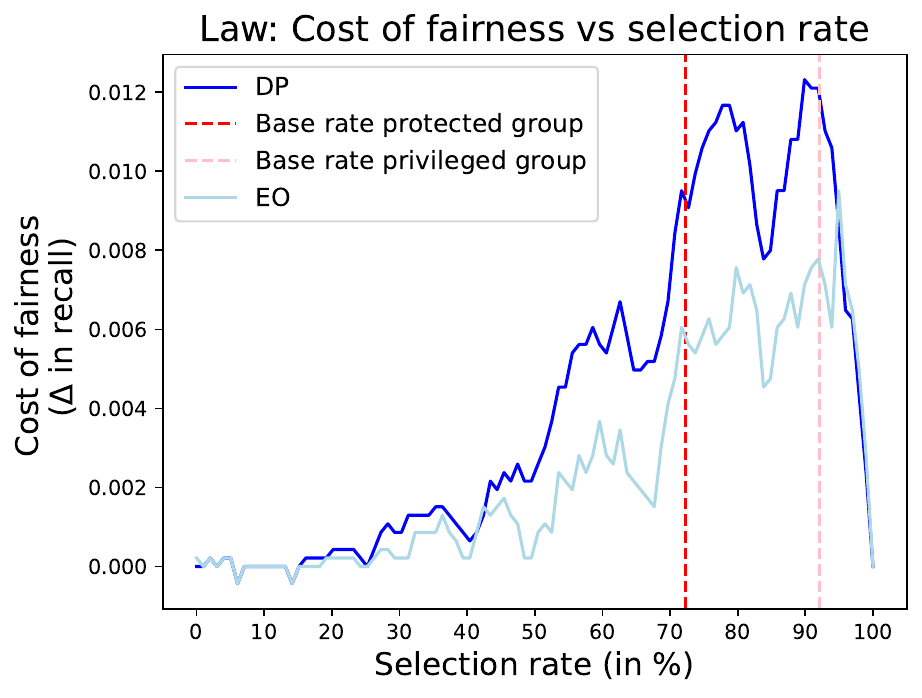}
            \caption{Law}
        \end{subfigure}
        \begin{subfigure}[b]{0.31\textwidth}
            \includegraphics[width=\textwidth]{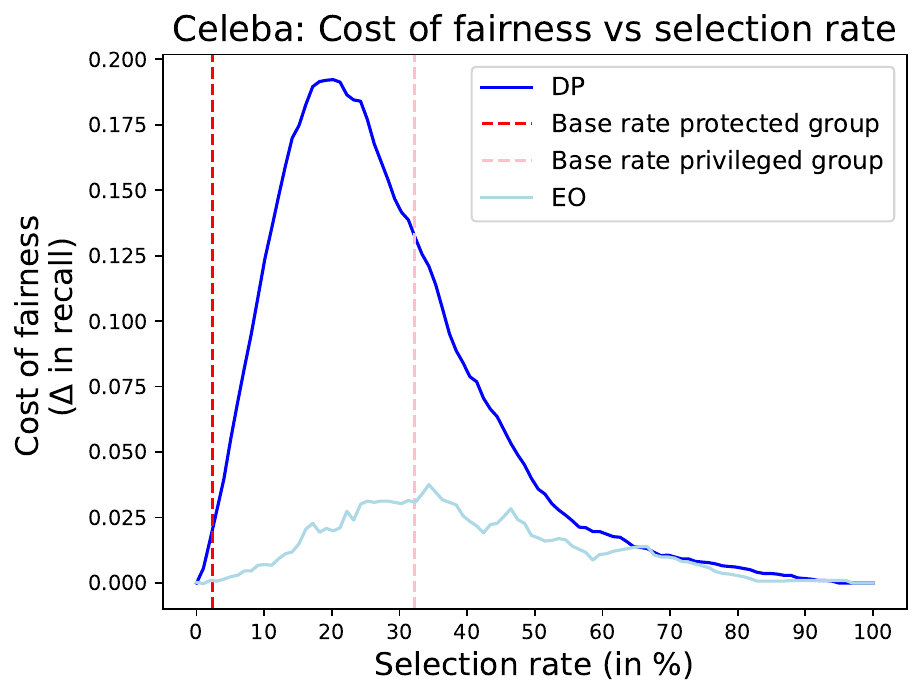}
            \caption{CelebA - Earrings}
        \end{subfigure}
        \begin{subfigure}[b]{0.31\textwidth}
            \includegraphics[width=\textwidth]{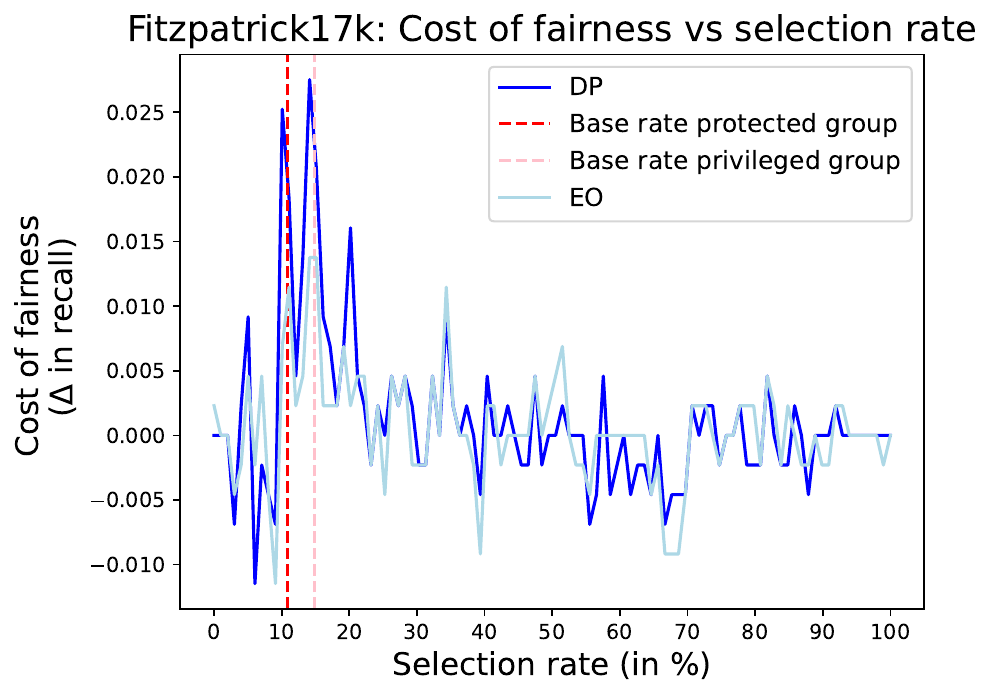}
            \caption{Fitzpatrick17k}
        \end{subfigure}
    \caption{Cost of fairness (loss in \textbf{recall}) for all datasets when enforcing demographic parity (DP) and equal opportunity (EO). The cost of fairness for both metrics heavily depends on the available resource level and thus the used selection rate.}
\label{fig:datasets_loss_recall}
\end{figure*}

While precision at the top $R$ relates to the efficient use of limited resources, knowing this value also determines the values of both recall and accuracy at $R$ for a specific dataset and model and a fixed value of $R$~\cite{rodolfa2021empirical}.
We measure the cost of fairness as the loss in precision when using the fair allocation instead of the optimal allocation.
For the sake of completeness, we also report the cost as loss in recall and loss in accuracy 
in Table~\ref{tab:model_perf} and Figure~\ref{fig:datasets_loss_recall}.
The formula to calculate recall at the top R:
\begin{equation*}
    \text{Recall at $R$} = \frac{\text{Number of actual positives in the top $R$}}{\text{Total number of actual positives}}
\end{equation*}
The formula to calculate accuracy at the top R:
\begin{equation*}
    \text{Accuracy at $R$} = \frac{\text{Number of actual positives in the top $R$} + \text{Number of actual negatives not in the top $R$}}{\text{Total population}}
\end{equation*}
We focus on these widely used performance metrics, but the results hold for any weighted sum of costs of Type I/II errors.

\subsubsection{Fairness Metrics}
We focus on two standard fairness measures used to assess disparities between groups. 

\textbf{Demographic parity} (also known as statistical parity) requires that the rate of positive decisions is roughly equal for both the disadvantaged group and the advantaged group~\cite{dwork2012fairness}: 
\begin{equation*}
        P(\hat{Y} = 1 | S = s) \approx P(\hat{Y} = 1 | S = ns)
\end{equation*}

\textbf{Equal opportunity} requires the true positive rate to be approximately the same across groups~\cite{hardt2016equality}, which enforces equal recall:
\begin{equation*}
        P(\hat{Y} = 1 | S = s, Y = 1) \approx P(\hat{Y} = 1 | S = ns, Y = 1)
\end{equation*}

\newpage

\section{Results for other performance metrics}
In this Section, we present the results for recall and accuracy. The general trends are in line with the results for precision.

\begin{figure*}[htb]
    \centering
        \begin{subfigure}[b]{0.31\textwidth}
            \includegraphics[width=\textwidth]{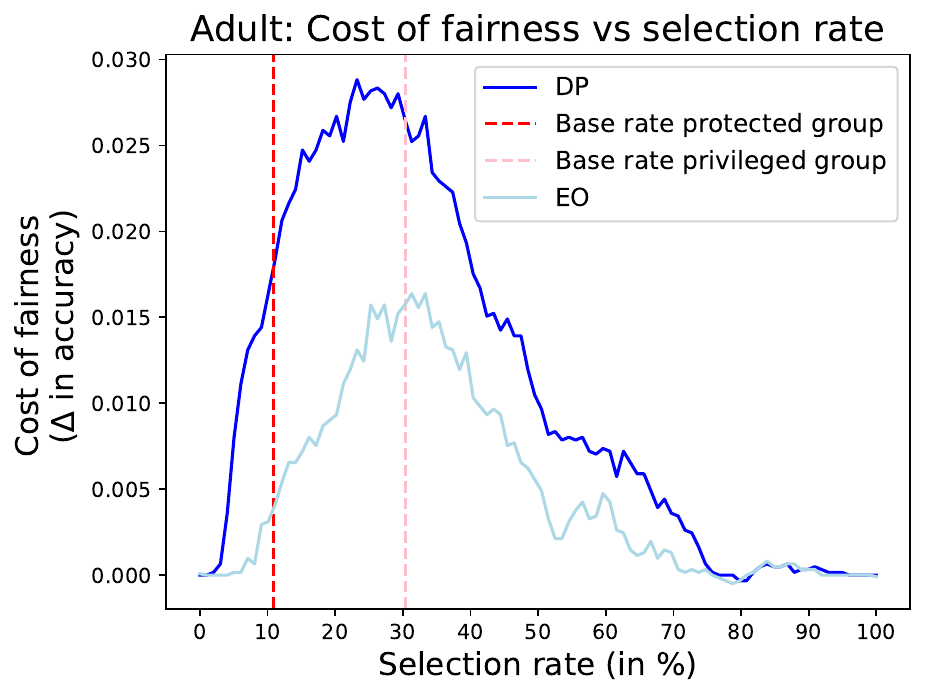}
            \caption{Adult}
        \end{subfigure}
        \begin{subfigure}[b]{0.31\textwidth}
            \includegraphics[width=\textwidth]{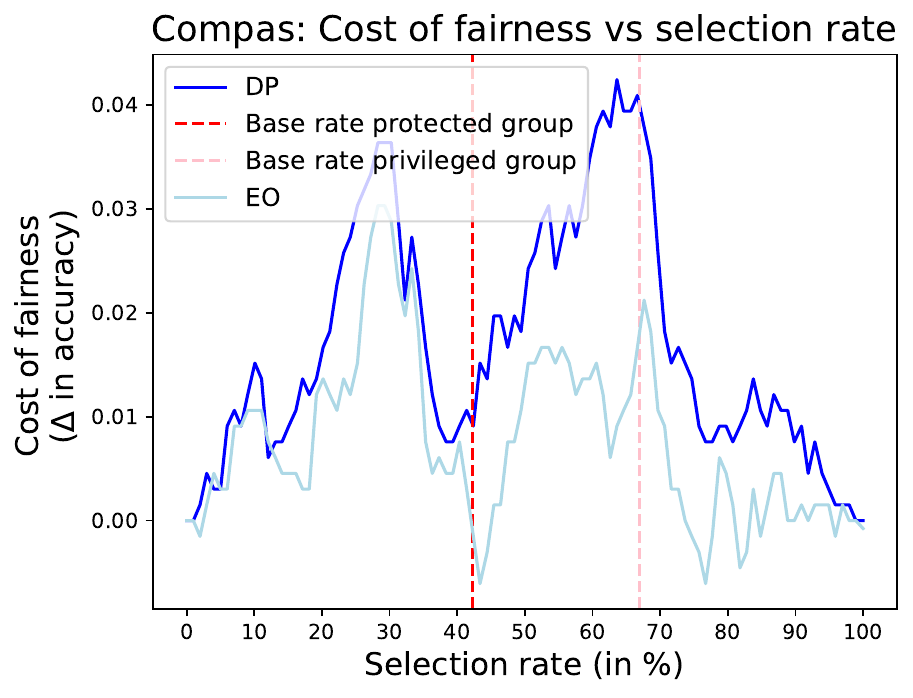}
            \caption{Compas}
        \end{subfigure}
        \begin{subfigure}[b]{0.31\textwidth}
            \includegraphics[width=\textwidth]{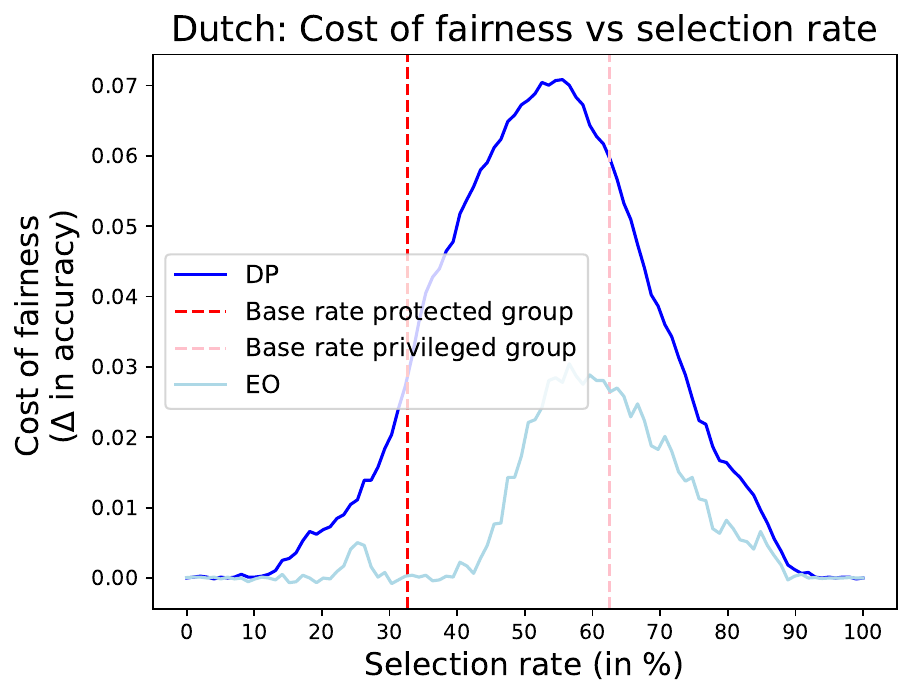}
            \caption{Dutch}
        \end{subfigure}
        \begin{subfigure}[b]{0.31\textwidth}
            \includegraphics[width=\textwidth]{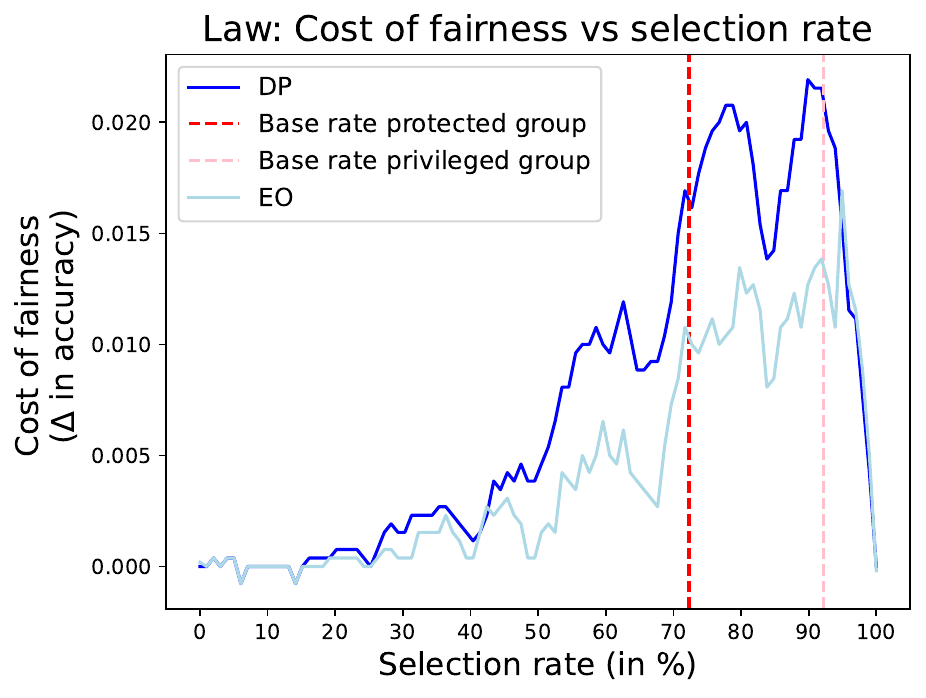}
            \caption{Law}
        \end{subfigure}
        \begin{subfigure}[b]{0.31\textwidth}
            \includegraphics[width=\textwidth]{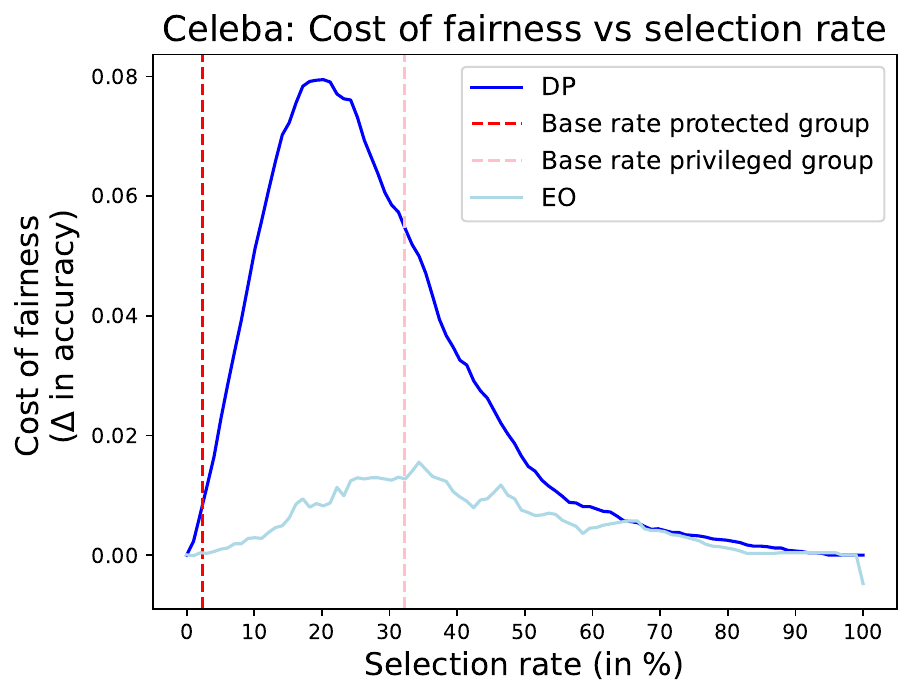}
            \caption{CelebA - Earrings}
        \end{subfigure}
        \begin{subfigure}[b]{0.31\textwidth}
            \includegraphics[width=\textwidth]{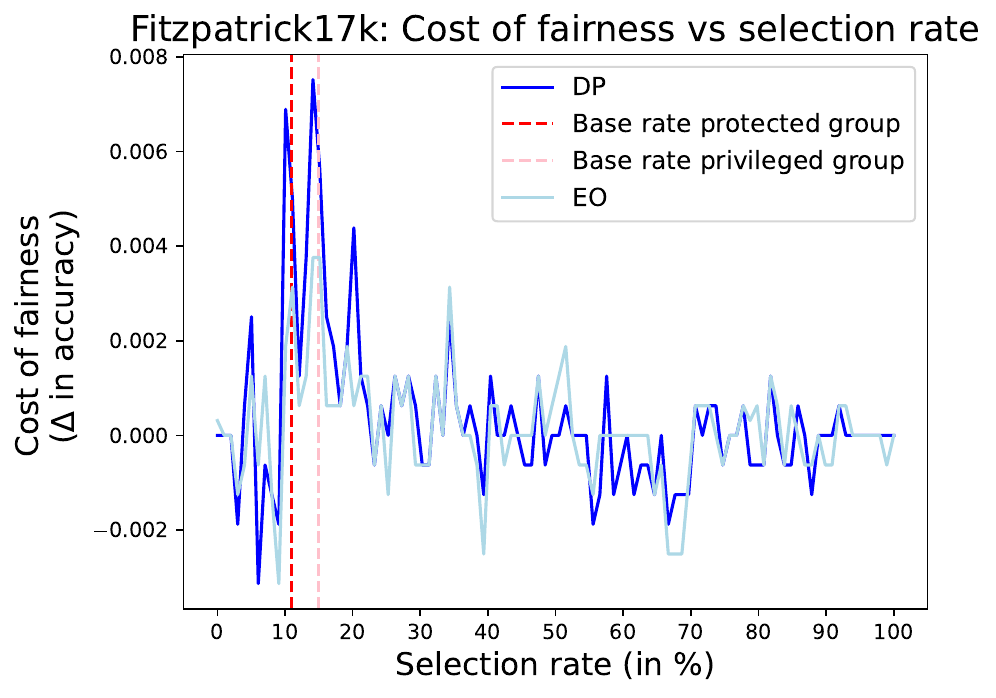}
            \caption{Fitzpatrick17k}
        \end{subfigure}
    \caption{Cost of fairness (loss in \textbf{accuracy}) for all datasets when enforcing demographic parity (DP) and equal opportunity (EO). The cost of fairness for both metrics heavily depends on the available resource level and thus the used selection rate.}
\label{fig:datasets_loss_accuracy}
\end{figure*}

\begin{table*}[htb]
\centering
\caption{The average cost of fairness (the loss in recall and accuracy) when enforcing DP and EO. The average is calculated over all the selection rates from $1\%$ to $100\%$.}
\label{tab:model_perf_recacc}
\begin{tabular}{c|cccccc}
Dataset& Adult& Compas& Dutch&Law & CelebA & Fitzpatrick17K \\ \hline
Avg. loss in recall (DP) & 0.023 & 0.016 & 0.028 & 0.004 & 0.062 & 0.001 \\
 Avg. loss in recall (EO) & 0.010 & 0.007 & 0.008& 0.002 & 0.013 & 0.001\\
 Avg. loss in accuracy (DP) & 0.011 & 0.017 & 0.026 & 0.007 & 0.026 & 0.000 \\
 Avg. loss in accuracy (EO) & 0.005 & 0.008 & 0.008& 0.004 & 0.005 & 0.000\\
 \end{tabular}
\end{table*}

\clearpage
\newpage
\section{Results for different resource levels} \label{sec:resource_levels}

\begin{figure}[htb]
    \centering
        \begin{subfigure}[b]{0.46\textwidth}
            \includegraphics[width=\textwidth]{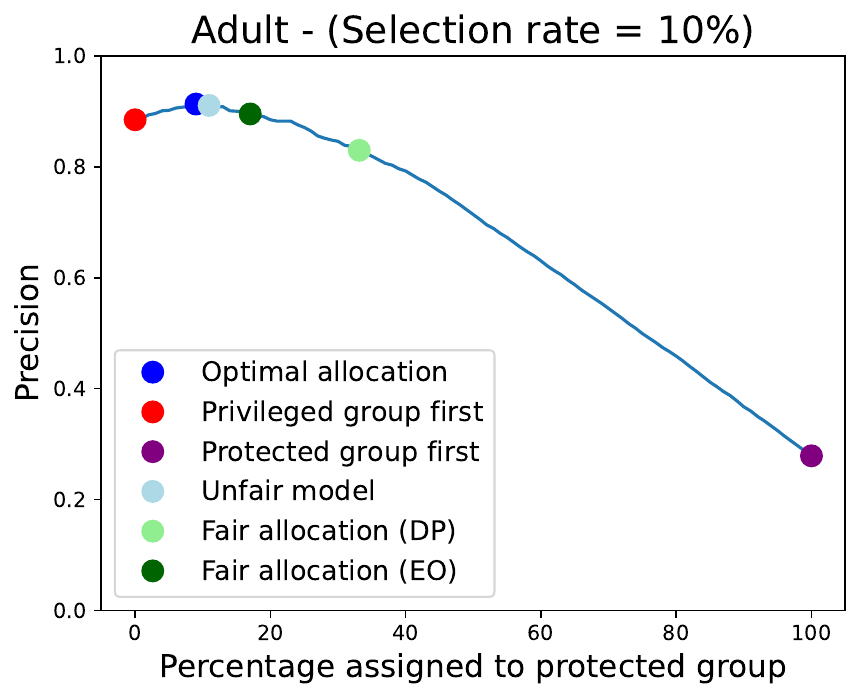}
            \caption{Selection rate = 10 \%}
            \label{fig:adult_rl10}
        \end{subfigure}
        \hfill 
        \begin{subfigure}[b]{0.46\textwidth}
            \includegraphics[width=\textwidth]{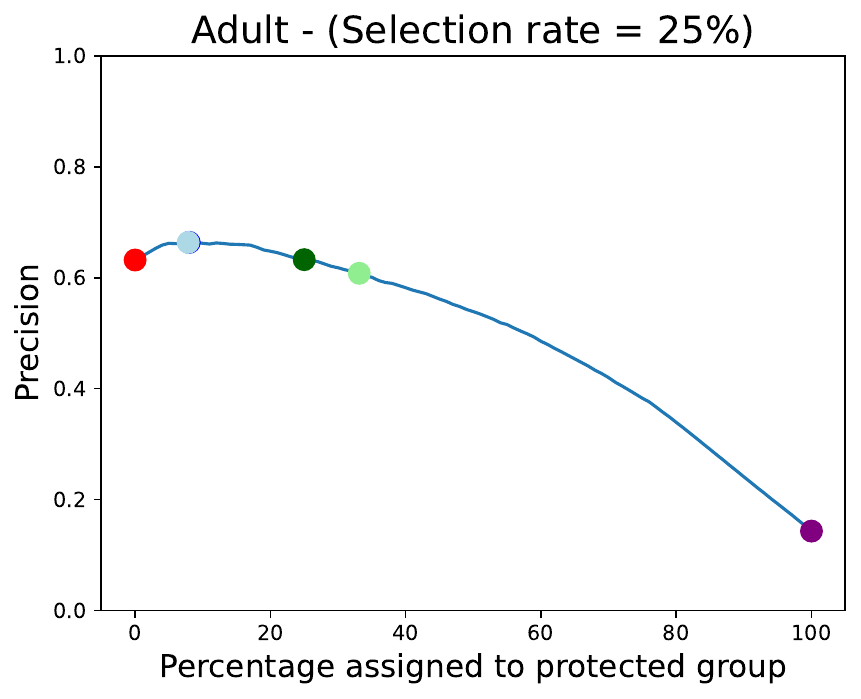}
            \caption{Selection rate = 25 \%}
            \label{fig:adult_rl25}
        \end{subfigure}

        \vspace{1em} 

        \begin{subfigure}[b]{0.46\textwidth}
            \includegraphics[width=\textwidth]{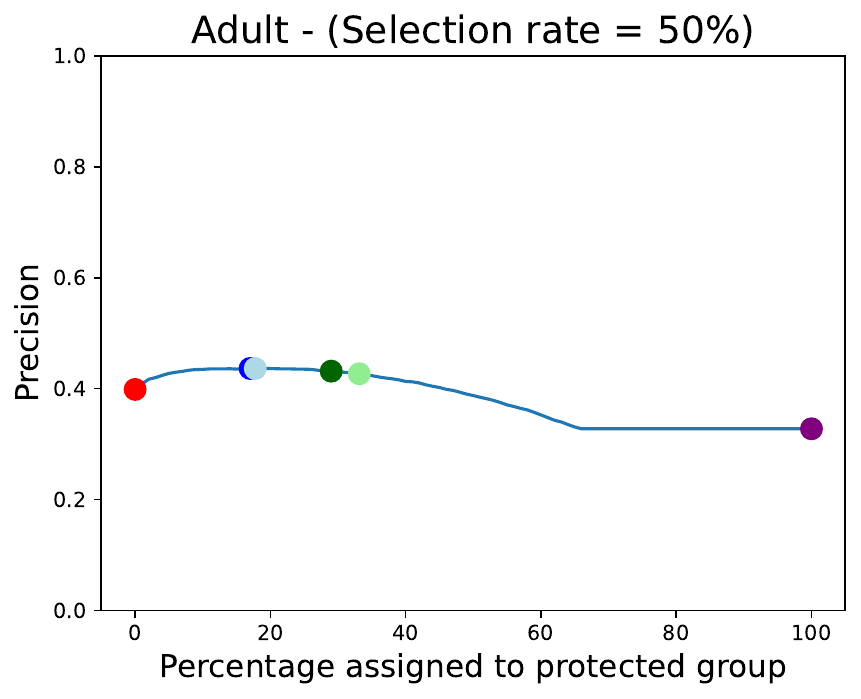}
            \caption{Selection rate = 50 \%}
            \label{fig:adult_rl50}
        \end{subfigure}
        \hfill 
        \begin{subfigure}[b]{0.46\textwidth}
            \includegraphics[width=\textwidth]{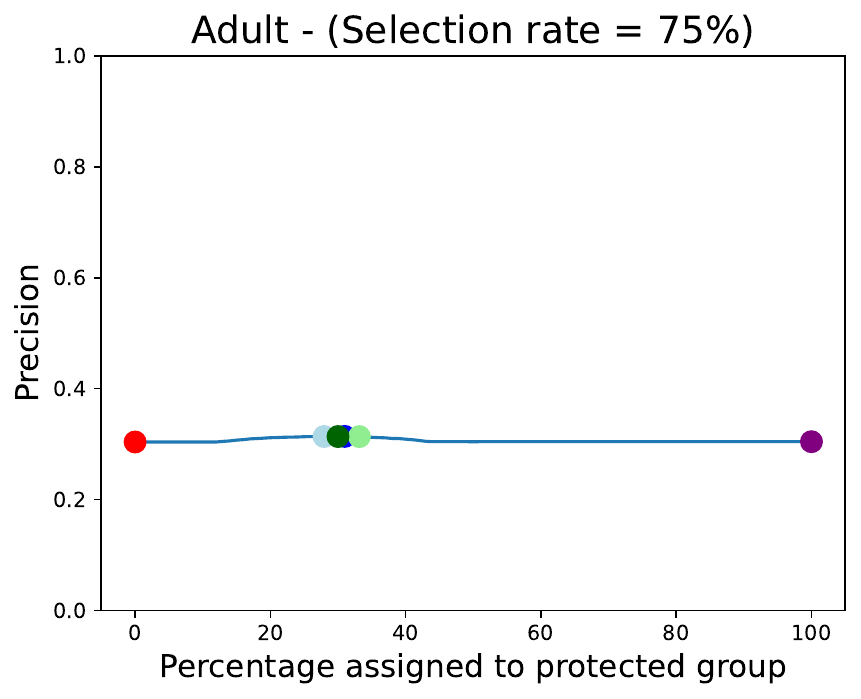}
            \caption{Selection rate  = 75 \%}
            \label{fig:adult_rl75}
        \end{subfigure}
    \caption{Allocation results for different resource levels (Adult Income dataset). Left point on the x-axis represents the resources being maximally allocated to the advantaged group, while the right point represents the resources being maximally allocated to the disadvantaged group.}
    \label{fig:adult_resourcelevels}
\end{figure}

We study the effect of the different resource level on the allocation of resources.
For the Adult dataset, this is shown in Figures~\ref{fig:adult_rl10}-\ref{fig:adult_rl75}.
These figures demonstrate how different allocations lead to different levels of precision for one fixed resource level. The two extreme points of the curve represent all the resources being awarded to either the advantaged group (red dot on the left) or the disadvantaged group (purple dot on the right).\footnote{However, also for these extreme points, when the total level of resources is higher than the size of one of the groups, some of the resources will still be awarded to the other group (as can for example be seen in Figure~\ref{fig:adult_rl50}).}
We see that the optimum is not reached by awarding all the resources to one of the groups, but somewhere in the middle~(darkblue dot).
We note that the precision of the unconstrained (`\emph{unfair}') model, will be very close to the optimal allocation, and that the allocations required by the fairness metrics (both DP and EO) will lead to a lower precision for every resource level (Figures~\ref{fig:adult_rl10}-\ref{fig:adult_rl75}). However, this difference is a lot larger for low resource levels (Figures~\ref{fig:adult_rl10}-\ref{fig:adult_rl25}). In Figure~\ref{fig:adult_rl50}, the default allocation of the machine learning model is very unfair, but using the fair allocation (both DP and EO) results in only a little change in precision. This demonstrates that a very unfair initial model allocation does not guarantee a high cost of fairness.
In Figure~\ref{fig:adult_rl75}, the default allocation of the machine learning model is already approximately fair. Hence, the cost of fairness is also low.

\end{document}